# Contextual Graph Representations for Task-driven 3D Perception and Planning

by

Christopher Agia

Supervisor: Prof. Florian Shkurti

April 2021

**B.A.Sc. Thesis**

Division of Engineering Science
UNIVERSITY OF TORONTO

***Abstract***— There is a growing demand for robot systems that are capable of executing compositional tasks in household environments with efficiency and robustness. Pure policy solutions offer efficiency by conditioning actions only on the current state; however, they struggle to generalize to long-horizons tasks. An appropriate strategy to deal with the increasing horizon of such tasks is to deduce a series of high-level, more achievable subgoals for the robot to accomplish. This process, known as task planning, is dependent on a suitable state representation that can be altered through high-level actions in order to achieve a goal.

Recent advances in the vision community facilitate fully automatic extraction of object-centric relational representations from visual-inertial data. These state representations, dubbed 3D Scene Graphs (SGs), are a hierarchical decomposition of real-world scenes with a dense multiplex graph structure. While SGs claim to promote efficient planning, they contain numerous objects and relations when only small subsets are required for a given task. This magnifies the state space that planners must operate over, and prohibits deployment in resource constrained settings.

In this thesis, we test the suitability of existing embodied AI environments for research at the intersection of task planning and 3D scene graphs, and construct a benchmark for empirical comparison of state-of-the-art classical planners. Furthermore, we explore the use of Graph Neural Networks to harness invariances in the relational structure of planning domains and learn representations that afford faster planning.

***Acknowledgements***— I would like to begin by thanking my advisor, Professor Florian Shkurti, for his utmost support and encouragement over the past year. It is very rare to meet someone who inspires you with their intellect, drive, and curiosity, and at the same time humbles you with their kindness, understanding, and selflessness. Most would agree that these qualities fall short of describing Florian as both a researcher and a friend.

I appreciate the mentorship of Professors Hani Naguib and Goldie Nejat. I have been so fortunate to meet Hani early on in my undergraduate years; he has treated me like family since. Likewise, the many insightful discussions shared with Goldie continue to influence my career decisions today.

To the colleagues, mentors, and friends that I have found in Krishna Murthy, Manfred Diaz, and Ran Cheng, thank you. They have been the best role models that I could ask for. Moreover, I would like to acknowledge Krishna's efforts and contributions to this project.

I have learned so much from working alongside Professors Dave Meger, Liam Paull, and Gregory Dudek; they have welcomed me with open arms on a number of collaborations. I would like to express my gratitude for their dedication, trust, and kind words of motivation.

Thank you to my dearest friends, Steve Kim, Christopher Mazzuca, Samuel Atkins, George Papangelakis, Samuel Looper, Svet Leniuk, and Anna Deza. Their friendship has brought color and joy to our shared journey through Engineering Science.

Lastly, I am grateful for my family, Remon, Sandra, and Emily, for their unconditional love, guidance, and light that they bring to my life.

To Emily

# Table of Contents

# List of Figures

# List of Tables

# Section 1: Introduction

There is a growing demand for intelligent robots that are capable of performing compositional tasks involving high-level reasoning, low-level motion planning, and control. If intended to be deployed alongside humans in industrial or household settings, these systems must operate safely, reliably, and with minimal latency by: (a) ensuring that the various independent processes are tightly integrated; (b) constructing representations of our environments that facilitate better planning and action. While much of the challenges to the former can be addressed through Integrated Task and Motion Planning [1] (TAMP), for which there exists a rich and growing body of literature, the robot vision and planning communities have yet to converge on a set of strategies to address the latter. The objective of our research is to incorporate *learning* over structured scene representations to make them more suitable for robot task planning.

An essential aspect of the planning problem is the *domain* or *map* upon which a plan is executed. Maps are often designed to include only features relevant to a specific task – for instance, navigation maps – which reduces the memory requirements and computational cost of planning over the map [2, 3]. While these maps are highly useful within the scope of their design, they are unable to provide the necessary information for a broader range of tasks. More recently, researchers have proposed the use of hierarchical scene graphs [4, 5] which describe multi-relational properties between scene entities in a layered structure, i.e., multiplex graphs [6]. The benefit of scene graphs lies in their ability to extend the planning challenge to a set of *long-horizon* and possibly *nested* tasks, where the final objective can be decomposed into the completion of successive sub-tasks. However, this extensibility trades off the speed of traditional planners as they must now search over a much larger state-space – even the most performative of planners can take on the order of a hundred seconds to find a solution for a given query [7]. Much of the recent literature has focused on developing improved heuristic-based planners [8, 9] and action-grounding algorithms [10], or attempting to learn a suitable policy from demonstrations [11]. Notable exceptions are [7], [12], and [13] which take a learning-to-plan approach for low-complexity and independent problem instances. Unfortunately, drawing admissible heuristics becomes increasingly difficult in large-scale graphs, and action-grounding / policy learning algorithms still struggle to generalize in small-scale single task domains [7].

Since only a small subset of entities, properties and relationships of the scene graph are considered for a single task, we hypothesize that learning a task-centric graph representation can better contribute to efficient task planning and policy learning. With modern advancements in graph representation learning, these planner-agnostic graph representations can be extracted with Graph Neural Networks (GNNs) [14]; a useful tool for processing graph features (locally irregular) with minimal memory and computational cost. These learned representations can greatly reduce the search space over which planning algorithms operate, facilitating real-time resource constrained planning. Furthermore, we aim to build a comprehensive benchmark of tasks in realistic domains to evaluate classical and learning techniques for task planning. The robotics community has historically treated perception as disjoint from planning and control. Success in this project could further influence the adoption of a more holistic view on perception, planning, and control, as well as motivate ongoing research at this intersection.

# Section 2: Literature Review

## 2.1 Background

For several decades, the robotics research community has pushed on the development of algorithms that enable robots to perform physical tasks requiring some degree of *intelligence*. Repetitive tasks in well-constrained domains, such as industrial robotic applications and advanced manufacturing, were quick to benefit from the speed and robustness provided by robotics systems at scale. However, while the ordinary person would not hesitate to carry out *simple* tasks in household or office settings, like *pour a cup of coffee*, it turns out that these tasks have presented the greatest of challenges to roboticists for several reasons. On a high-level it requires, amongst other things: operation in partially unknown / unconstrained settings (localization and mapping), extracting the right features from sensors in the presence of observation noise (perception), constructing efficient trajectories for robot motion and precisely acting them out (motion planning and control), and handling stochasticity in environment interactions (modelling). In the design of robotic systems, the objective is to bring these processes together to handle task execution in settings that cannot be known beforehand. Moreover, extending robot capabilities to execute arbitrary tasks of increasing length and complexity requires *planning*.

Notice that the above example, pour a cup of coffee, can be decomposed into a sequence of smaller, more achievable tasks: grab the coffee pot, pour the coffee, and place the coffee pot – such low-level tasks can be accomplished with single motion primitives. Efficient high-level reasoning over potential states and actions in order to simplify a given task is an essential characteristic of an intelligent agent, as providing step-by-step instructions becomes impractical for long compositional tasks like *clean the room*, and insufficient in different environments.

Robot task planning and symbolic reasoning algorithms typically operate over a pre-defined data structure composed of *symbols* or *entities* whose properties are altered by the algorithm with actions to accomplish a task. Naturally, graphs of various types are used to define the planning domain, with the finite state space represented by nodes, and action-conditioned state transitions represented by edges. Certain graph properties may deter the performance of search algorithms, in some cases producing search inefficiencies or incompleteness, e.g., depth first search on cyclic graphs. This begs the question: what type of graphs are most suitable for robot task planning?

Ideally, we require a graph that: (a) contains all interactable scene entities; (b) expresses useful relationships between them in an arrangement or grouping that accurately reflects the physics of the scene; (c) provides a dense metric representation of the scene for integrated task and motion planning. Also, automatic construction of the graph from high-dimensional observations such as images or videos is important, as it would enable *replanning* in regions not previously explored. Pertaining to these criteria, researchers have proposed the use of hierarchical scene graphs [4, 5, 15] to abstract high-dimensional scene information into a structured multi-layered graph. Layers of the graph correspond to pre-defined levels of abstraction: metric-semantic level, objects, places, rooms, and buildings. Nodes represent the class and properties of scene entities at a particular level, and edges describe multi-relational properties between the scene entities. Scene graphs are discussed in further detail in Section 2.2.

This thesis investigates the feasibility of robot task planning in large-scale 3D scene graphs, comprised of several orders of magnitude more elements than experimented with in prior works [7, 13, 16, 17]. Furthermore, we are interested in long-horizon tasks as opposed to object search or retrieval, which introduces a fundamental distinction between the *scene graph* and the *planning graph*. In particular, our tasks involve altering the unary and pairwise properties of entities within the scene graph in a sequence of ground actions – this could be characterized as a form of *rearrangement* [18]. This formulation equates a node in the planning graph to an assignment of properties and predicates to all entities in the scene graph. With majority of robot actions being reversable, such as pick-up and drop, the resultant planning graph has a directed cyclic structure. In addition, the vast number of scene entities and possible actions yields a combinatorial explosion on the graphs branching factor. This poses a significant challenge for classical planning techniques that do not leverage the hierarchical structure of the scene graph or the underlying distribution of objects when computing heuristics, and learning-based methods that have yet to generalize across tasks and scenes at a practical scale.

Different planning approaches offer unique advantageous and drawbacks. Upcoming sections will cover prominent works relevant to robot task planning across two main categories:

- Classical Planning [19, 20, 21] - Section 2.3: A collection of planning methods that aim to derive heuristics from relaxed problem instances to guide their search. These methods are domain-agnostic and interpretable, but slow in large problem instances and do not improve over time. These off-the-shelf methods can be applied to any planning problem casted into standard planning language format.
- Learning to Plan [16, 17, 22] - Section 2.4: Modern planning techniques that focus on leveraging data of past queries or *expert demonstrations* to learn generalizable heuristics. These learning-based approaches can yield efficient search and are more flexible than their classical planning counterparts. However, they are data hungry, computationally demanding, domain-dependent, and lack interpretability. These methods also require training, which could be a time-consuming process.

There is also an emerging stream of hybrid methods, which aim to combine benefits from classical methods (theoretical guarantees, search styles) and powerful artificial intelligence tools (function approximation, optimization) [13, 23]. We believe that optimized planning over scene graphs should inherit the strengths of both approaches; the design of an efficient and generalizable approach to robot task planning in large-scale scene graphs is one of our main objectives.

### 2.1.1 Definition of a Planning Problem

We follow the formal definition of a planning problem as defined by *Silver and Chitnis et al.* [7]: a tuple $\psi = (\mathcal{P}, \mathcal{A}, \mathcal{T}, \mathcal{O}, \mathcal{I}, \mathcal{G})$ encapsulates a finite set of properties and predicates ($\mathcal{P}$), actions ($\mathcal{A}$), and objects ($\mathcal{O}$). Let the finite set of all possible states be denoted with $\mathcal{S}$. As mentioned previously, a state $s \in \mathcal{S}$ is an assignment of values to all properties over objects and corresponds to a single node in the planning graph. Properties are real-valued functions over a set of objects with an arity that typically does not exceed three. For example, a pairwise property expressing the percentage area of *cup3* visible behind *plate1* could be *co-visible(plate1, cup3) = 0.6*. Predicates or literals are a type of property that are binary-valued; for instance, *is-cup(cup3) = true* or *is-contact(cup3, plate1) = false*. The initial and goal states of the scene graph are denoted by $\mathcal{I}$ and $\mathcal{G}$, respectively. We note that the goal state $\mathcal{G}$ need not assign values to all properties, but rather, can impose requirements to a property subset $\hat{\mathcal{P}} \subseteq \mathcal{P}$ over an object subset $\hat{\mathcal{O}} \subseteq \mathcal{O}$. This is referred to as the closed world assumption (CWA); all facts not listed in the goal condition are assumed to be false. Actions are object-parameterized functions that are defined as *grounded* when applied over specific object instances. For example, *pickup(?x)* becomes a ground action *pickup(cup3)* when applied over object *cup3*. The state transition function, $\mathcal{T}$, reflects the environment dynamics by mapping a state and action to a next state with probability $p$. A plan is any sequence of actions that drives the state of the scene graph from $\mathcal{I}$ to $\mathcal{G}$. Lastly, the goal of a planner is to find a solution that maximizes the rewards (or minimizes the cost) with respect to an objective function, which could be as simple as the length of a computed solution.

With a focus on high-level task planning, we mainly consider deterministic transitions between states because auxiliary robot subsystems (e.g., vision, motion planning, and control) are assumed to be solved, and the environment is static. However, potential extensions to stochastic environments with dynamic agents is of interest to us in future work.

### 2.1.2 The Planning Domain Definition Language

Rapid development in algorithmic planners in the late 20th century warranted a standardized problem specification language that extended the STRIPS encoding to include a variety of features to describe increasingly complex problems. Namely, negated preconditions, type specification of objects, and numeric variables. The Planning Domain Definition Language (PDDL) [24, 25] became standard at the Artificial Intelligence Planning Systems Competition (AIPS-98) [26], and has since been extended to account for stochastic environments [27], continuous and discrete dynamics models [28], and black-box sampling processes [29]. The latter methods are used for TAMP research; the standard PDDL framework is sufficient for symbolic reasoning over static scene graphs.

PDDL is used to instantiate the planning problem as defined in Section 2.1.1 with a domain file and a problem file. The domain file outlines the set of entity types, predicates, object-parameterized actions, and constants which are domain specific and will be used for all planning problems. The problem file is used to realize a specific problem instance by defining the set of objects in the domain, assigning initial values to predicates over all objects, i.e., the initial state, and specifying the goal condition. An action can be executed at a state if the state satisfies the actions *preconditions*, and the *effect* of the action transitions the current state to the next. PDDL is also the supporting language in several robotics benchmark simulators [11, 30]. We have selected an extension to the ALFRED home simulator as a platform for preliminary experiments because it provides tooling to cast a generated task and scene graph into PDDL format to be solved by a classical planner.

### 2.1.3 Rearrangement: A Challenge for Embodied AI

The rate of development in the computer vision and natural language processing (NLP) communities has been resounding. Facilitated by popular benchmarks with widely accepted base task specifications and standardized evaluation metrics, these fields have benefited from breakthrough machine learning algorithms and architectures that have accelerated research in these communities as a whole. Unlike vision and NLP, research in Embodied AI – the study of intelligent systems with a physical or virtual embodiment – has lacked consistency in the types of tasks that said *embodied AI* agents are designed to perform. As such, many newly proposed methods are being analyzed with different sets of evaluation metrics, on physical embodiments (robotic systems) that depend on varying sensor suites, and in domains that present different challenges for the embodied agents. These inconsistencies make it nearly impossible to fairly compare methods. Moreover, it slows the rate of progress in the research community, as each new effort must assume the overhead of adapting and benchmarking prior works for side-by-side comparison. Unfortunately, the solution is not as simple as taking a one-simulator-fits-all approach; the release of new and improved simulators is inevitable and necessary, as each is equipped with a distinct set of features for experimenting and developing unique robot capabilities (e.g., assembly versus navigation). Varying levels of granularity in the modelling of physical interactions, stochasticity and dynamicity of the environment, and diversity in scenes and objects are but few ways in which modern simulators differ. Therefore, at a high-level, the objective of the Rearrangement framework [18] is to reconcile the demand for a common task specification, and consistent testing and evaluation protocols, with the vast differences between the popular existing experimental platforms and the up-and-coming simulators of the future.

Primarily, Rearrangement describes a superclass the encapsulates the majority of tasks that embodied AI researchers seek to solve. That being, the ability to rearrange the current state of a system into a specified goal configuration. For instance, if the *state* is simply defined by the state variables of an articulated manipulator, then rearrangement describes a motion planning problem, where the system must compute a set of low-level motor forces and torques that drive the state variables into the goal configuration. In a more complicated setting, the *state* may refer to the pose and physical properties of all objects and agents in a scene, and the goal configuration could specify an alteration of various physical properties and the positions of such objects. In either case, it is essential that an intelligent agent could autonomously limit the scope of its physical interactions to the spaces or set of objects most related to the task at hand. For example, the articulated manipulator should not violate the constraints of its configuration space, and an intelligent mobile agent should not turn on the stove when searching for a pan. To account for this, Rearrangement proposes to incorporate a domain-specific *do no harm* test, which can be interpreted as a maximum threshold for redundant or potentially dangerous interactions enacted by an agent before the plan is invalidated as a whole.

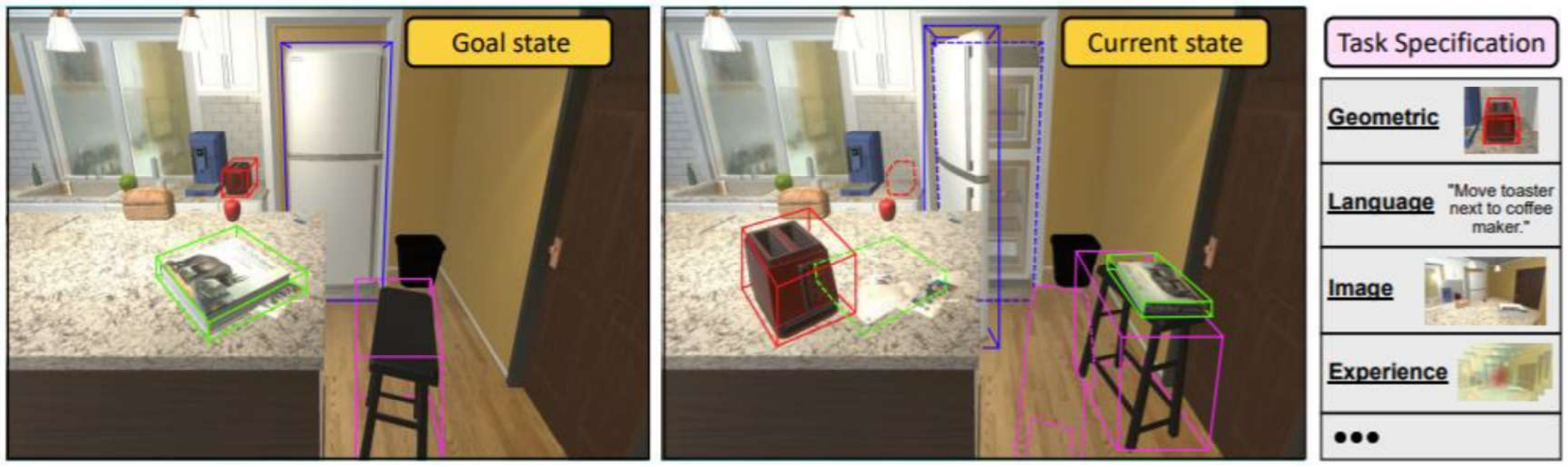

Figure 1: The task of rearrangement for embodied AI research [18].

The formal definition of the rearrangement task falls under the Partially Observable Markov Decision Process (POMDP) framework. While sharing some similarities to the definition of a planning problem as described in Section 2.1.1, the main distinction lies in that the agent must operate on observations (or emissions) $o_t \in \mathcal{O}$, as opposed to having full access to the underlying state of the environment $s \in \mathcal{S}$. This emphasizes the need for a robust perception pipeline that can construct actionable state representations $\hat{s_t} = \phi(o_t)$, where $\hat{s} \in \hat{\mathcal{S}}$. In particular, we define:

$$\hat{\mathcal{S}} = \{\hat{s} \mid \hat{s} = \phi(o), \forall o \in \mathcal{O}\}$$

as the set of all possible states constructed by a perception module $\phi(\cdot)$. In alignment with this definition, our research project identifies $\hat{\mathcal{S}}$ as the set of possible scene graphs states, and $\phi(\cdot)$ as a perception system that can construct scene graphs from high-dimensional sensory inputs. We seek to test the hypothesis that hierarchical 3D scene graphs are the right actionable representations for rearrangement-type tasks, and in doing so, we take much inspiration from this work in terms of their proposed metrics and guidelines. Certainly, approaching the full rearrangement task in photorealistic robot simulators with built-in physics engines [31, 32, 33, 34] would involve many subsystems such as real-time perception, motion planning, and localization, to name a few. However, the inclusion of these systems extend the scope of this project, and our choice to focus on the task planning component of Rearrangement is motivated by: (a) the fundamental characteristic of an intelligent robot to decompose long-horizon tasks into more achievable subgoals, e.g., to compute a sequence of high-level actions from a predicate-based specification of a scene graph, and (b) the well-established pitfalls of classical task planners when operating in large domains with many extraneous objects. As we will describe in the following section, scene graphs fit this narrative almost exactly, containing many objects and defining many relationships that are irrelevant for a specified task. This produces inefficiencies in planning algorithms which very seldom leverage learning to reason over the important parts of the state. In summary, we adopt the Rearrangement framework as a basis for our robot task planning research in 3D scene graphs and potential extensions to motion planning.

## 2.2 Spatial Perception to 3D Scene Graphs

Robot planning requires symbols grounded in a space that best captures the properties and physics of a scene. In previous years, the computer vision community has focused on grounding semantic information in the space of images [35, 36]. However, image pixels are subject to high variance with the smallest of parameter changes and are not invariant to different perspectives. Images also fail to capture the full 3D geometry of the scene; at best, partial geometry can be captured with depth information drawn from sensors or determined through non-linear optimization of key-point correspondences. Thus, constructing viable long-horizon plans directly from images has had little success [11]. However, rich features found in images – commonly used in discriminative scene understanding tasks – could provide utility in combination with an abstract scene representation. Previous works in map design for robotics have focused on static 2D hierarchical representations [37, 38, 39] and non-hierarchical 3D metric-semantic maps [40, 41]. Unfortunately, these works do not provide the appropriate actionable abstractions to jointly facilitate high-level reasoning and motion planning.

Scene graphs (SG) have previously been used in computer graphics, but were recently brought to light for robotics applications by *Armeni et al.* [4] and *Kim et al.* [15]. While [15] misses several layers of abstraction in their model, [4] propose a four layered structure to describe the scene semantics, 3D space and cameras: *buildings, rooms, objects, cameras*, as shown in Figure 2. A fixed number of properties and relationships are encoded per node type, which can be interpreted as node and edge features of the graph. Examples of node properties include action affordance, location, size, mesh segmentations, and volume. Edges are used to define spatial or logical relationships between scene entities and are akin to properties with an arity larger than one in PDDL. A few examples are relative magnitude, occlusion, and parent space. The graph is constructed from panoramic images in a semi-automatic four step procedure leveraging performant semantic detectors [36]. Albeit a step in the right direction, human verification slows the graph generation process and could introduce inconsistency across scenes, creating an obstacle in the design of a general planning solution.

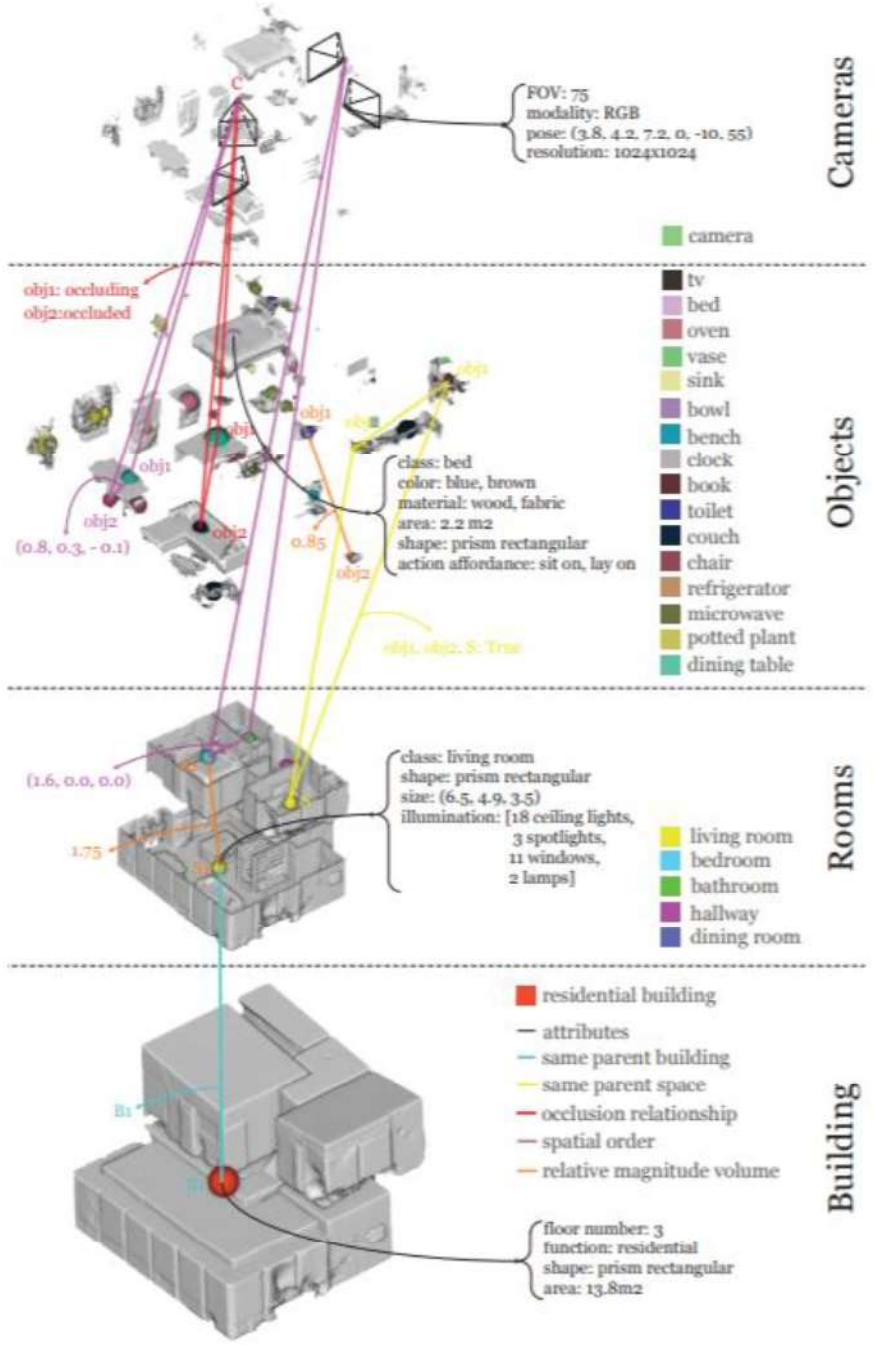

Figure 2: Levels of abstraction within 3D Scene Graph [4].

Dynamic Scene Graphs (DSGs) proposed by *Rosinol et al.* [5] addresses these concerns with a fully-automatic graph construction algorithm dubbed Spatial Perception Engine (SPIN), built off Kimera [42], which generates a 3D scene graph directly from visual-inertial data. The automatic construction allows for *consistent hierarchical abstractions* across dynamic scenes, superseding the limitations of prior map designs. Nodes and edges express many of the same properties and relationships as SGs, however, an extra layer for *places and structures* is incorporated between the *objects and agents* and *rooms* layer. In addition, DSGs dynamically model the shape and movement of agents in a scene with timestamped SMPL [43] models, which is a key enabler for TAMP research in crowded environments.

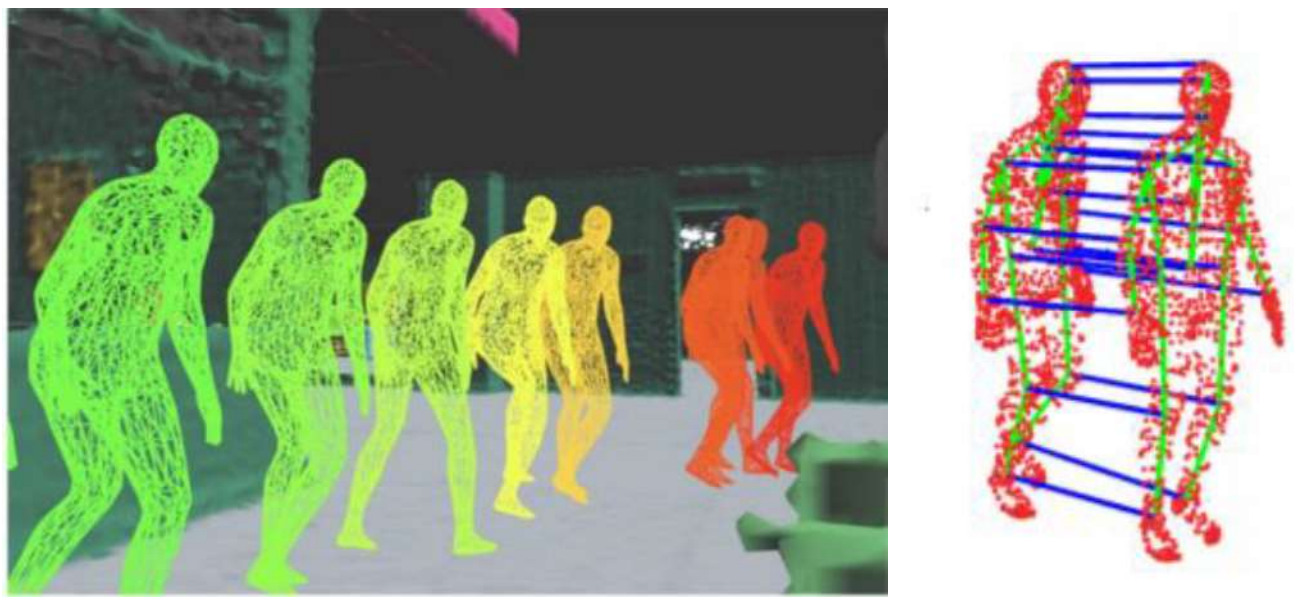

Figure 3: DSGs SMPL agent pose graph in the metric-semantic mesh [5].

As stated in [5], 3D Scene Graphs were designed as a flexible framework that **supports** TAMP queries with semantic node attributes for task planning from high-level task specification, and geometric node attributes for motion planning. However, SGs are not necessarily **optimized** for efficient planning in large compositional task settings – this is intuitive, since the graph contains all objects regardless of the query. As 3D scene graphs are being increasingly used in the development of robot simulators [11, 44], the necessity for research at the intersection of the spatial perception, scene graphs, planning and control is rapidly growing.

## 2.3 Classical Planning Techniques

In large and complex planning problems, traditional forwards and backwards search algorithms that guarantee completeness may need to visit millions or billions of states before arriving at solution. This results from the fact that the number of nodes visited in a search is exponential with respect to the average branching factor of the graph and the depth of solution. Hence, imposing operational constraints over these algorithms make them infeasible for deployment on autonomous systems that require real-time planning. Over the years, different classes of heuristics have been developed to more efficiently guide the search of planners to *satisficing* solutions [19, 20, 21], while other streams of classical planning such as SAT solvers have had success in generating *optimal* solutions [45, 46]. In robotics, we are often concerned with computing a reasonable plan at minimal cost (non-optimal sequential planning), rather than expensively searching for an optimal plan (optimal parallel planning). We herein focus on heuristics planning with ties to recent advances in the robotics planning community.

In heuristics planning, the goal is to craft a cost-to-go function from a general declarative problem specification (STRIPS or PDDL) that estimates a lower-bound on the distance from a queried state to a goal. Heuristics that satisfy this condition are termed *admissible*, and can be used to constrain the search of an algorithm to more promising paths of the graph. Methods for deriving heuristics come in many forms including abstraction-based planning [47], detecting landmarks [48], critical paths planning [49], and ignoring delete lists. Among these, ignoring delete lists is a generic simplification principle that has been incorporated in the design of state-of-the-art PDDL planners like Planning as Heuristic Search (HSP) [19], Fast Forward Plan [20] (FF-plan), and Fast Downward Plan [21] (FD-plan), each of which have been recently featured in robot task-planning works [7, 11]. In addition to their unique heuristics, another dimension in which they differ are their search strategy; the mechanism by which nodes are selected for expansion and the types of data structures that are used to store nodes on the search frontier. This is a critical design decision in the development of planning algorithms, as it heavily dictates the overall complexity and computational requirements of the planner. To list a few well-studied strategies, we have: Dijkstra's Algorithm, A*, IDA*, Greedy Best-First Search, Hill-climbing, Steepest-Ascent Hill-climbing, and Enforced Hill-climbing, amongst many others. Its not uncommon that given the same heuristic, multiple search strategies produce the same solution. However, this makes no implication on the similarity of their search, which can vary substantially in terms of number of visited states and the overall memory required for the search. To add another dimension of complexity, the theoretical guarantees offered by planning algorithms are derived from the combination of the heuristic and the search strategy, and hence, the design of one component is often coupled with the design of the other. For instance, A* guarantees optimality with an admissible cost-to-go, while hill-climbing strategies offer no such guarantee. As such, we will find that the ways in which classical methods attempt to maximize the speed-optimality trade-off is more often than not influenced by domain-specific knowledge, like the anticipated size of the state and action spaces. In addition, several of the aforementioned planners were designed to adhere to explicit computational constraints imposed by international competition rules, since they must all operate on identical hardware.

The heuristic defined by the HSP planner approximates the cost of the optimal (exponential-time) solution in the relaxed planning graph (RPG), where delete lists are ignored [50]. It is shown that this heuristic is inadmissible, and nonetheless, still very costly to compute. Its evaluation amounts to approximately 80% of the total runtime, despite using the crude hill-climbing search strategy to minimize the number of heuristic evaluations [19]. Although not optimal nor complete, HSP still maintains strong performance across a wide-range of classical planning problems [26] demonstrating the effectiveness of its heuristic function. *Bonet et al.* [51] proposed HSP2, swapping hill-climbing for a faster and more robust best first search strategy. HSP2 is further compared to a regression planning variant HSPr with a 6-7 times faster node expansion rate. As action preconditions are assumed to be independent in the HSP system, HSPr simply uses heuristics derived from atoms in the initial state without re-computation during the backwards search, accumulating only the costs of preconditions in new states. This substantial increase in speed comes at the cost of a lower quality heuristic and spurious states.

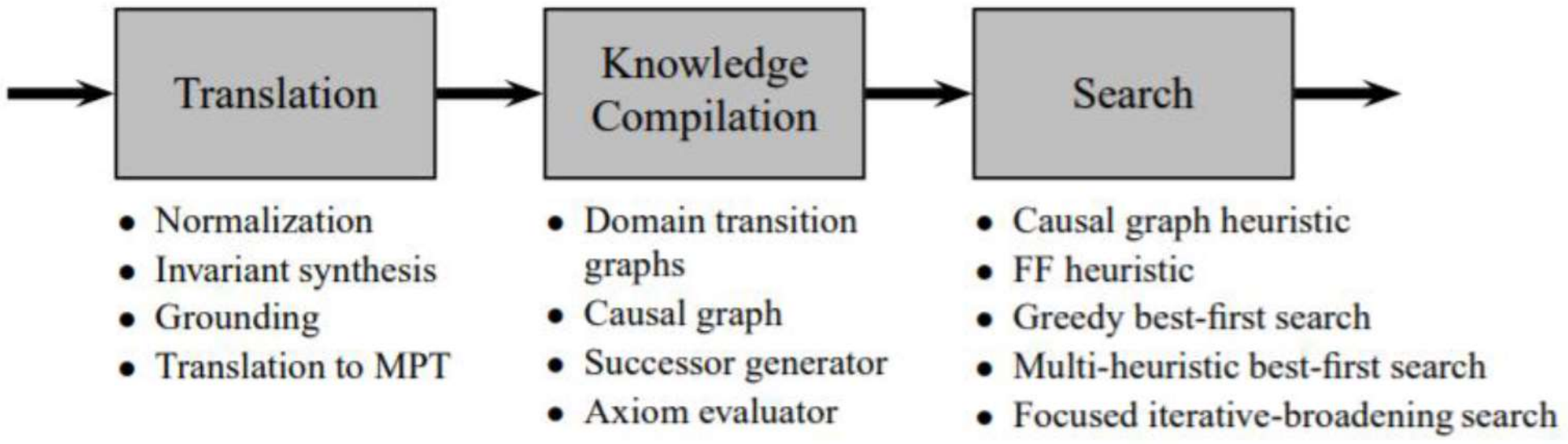

Figure 4: Phases of the fast-downward planning system [21].

The FF-plan system [20] makes several key improvements over HSP. Namely, it uses a more sophisticated heuristic evaluation function, alters the search strategy to enforced hill-climbing, and introduces action pruning. In particular, the heuristic evaluation leverages GraphPlan [52] to compute a polynomial-time solution to the RPG, accounting for positive interactions between predicates. Thus, heuristics computed by FF-plan are typically lower than those of HSP, while the cost of computing them are similar. The enforced hill-climbing strategy ensures that a solution will be found if the planning problem contains no dead-ends. However, this improvement does not suffice in scene graph planning, as the planning landscape contains a multitude of local minima that may cause this planner to fail. One simple example is the presence of irreversible action effects, such as slicing fruit or heating a meal. Helpful action sets are composed of actions that relate to at least one goal condition at the lowest level of the RPG. They can be built directly from the GraphPlan RPG solution, pruning redundant actions at each state. FF-plan is supplanted by FD-plan [21]; another heuristic progression planner that derives its heuristic from the inherent hierarchical structure of causal graphs. It employs three independent search algorithms, one of which is a multi-heuristic best first search combining the FF-plan heuristic with the causal graph heuristic in a complementary manner. The FD-plan system is capable of processing full PDDL2.2 problems [53], and outperforms HSP and FF-plan on most domains. Several notable works accounting for non-linear continuous dynamics state transitions (PDDL+) [28, 54] and BlackBox sampling procedures (PDDLStream) [29] have also shown strong results on a range of PDDL domains.

## 2.4 Modern Planning & Learning to Search

The general heuristics and search strategies that compose the planners presented in Section 2.3 make them applicable to a wide-range of problems, but they fall short on several accounts. For one, they are unable to improve with time or experience, and they fail to incorporate domain-specific priors. Instead, by increasing the degree in which planners are tailored to hypothesis set of problems, we can expect proportional improvements in their performance, and benefit from more optimal trajectories in shorter times [19]. *Learning to plan* algorithms seek to provide general planning frameworks that are differentiable, and hence, can be optimized for specific tasks through gradient-based methods. Learning over unstructured data such as scene graphs demands modern deep learning tooling, by which the likes of recent advances in the graph representation learning community are highly applicable. In the following section, we introduce key concepts of graph representation learning with specific emphasis on the class of approaches that operate over multi-relational data structures. These concepts will serve as a basis for the architectures and optimization schemes of contemporary learning to search methods discussed in Section 2.4.2.

### 2.4.1 Introduction to Graph Representation Learning

In recent years, deep neural networks have revolutionized the landscape of graph representation learning. There has been a drastic shift in state-of-the-art methodologies which have produced exciting results in fields ranging from molecular biology to social networking [55, 56]. Learning useful representations of non-Euclidean graph structures has been a long-standing classical challenge for several reasons: (a) the shear magnitude of graphs and their associated variances across domains; (b) the difficulties of hand-crafting generic approaches to detect graph properties or encode reliable / transferable embeddings; (c) the domain expertise required to effectively utilize classical techniques, along with an unforgiving tuning process. However, understanding the traditional graph representation learning approaches can provide valuable insight on the theoretical and practical motivations behind the newly popularized graph neural network models (GNNs). While an overview of classical graph learning techniques is out of the scope of this thesis, we refer the readers to *Hamilton et al.* [6, pp. 9-27] for a detailed summary.

The objective of graph representation learning is to be able to extract informative node and edge level features or *embeddings* from an input graph, in the hopes that the inductive biases learned from smaller training instances (often self-supervised) will generalize to larger and more complex problem instances. We use the term embedding to define a feature space for graph nodes and edges, as shown in Figure 4. Depending on the task, the embedding could encode local properties of the graph such as node semantics, local relationships, neighborhood clusters, or may capture global properties like the graph class, spanning structures, or path-based statistics. GNNs facilitate feature extraction from an input graph with fully-differentiable and highly parallelizable sparse matrix routines, superseding the use of *shallow embeddings* (i.e., look-up tables) which are *transductive*, do not allow for weight sharing, and fail to leverage valuable input features in the encoding phase [57]. The decoder, which need not necessarily be a GNN, transforms the embeddings into an output that is conducive to a particular task and objective function, thus allowing for the network parameters to be updated via backpropagation. Similar to most learning-based systems, the GNN architecture and objective function used to train it must be selected to best suit the graph type, such homogenous or multi-relational graphs, and the desired task.

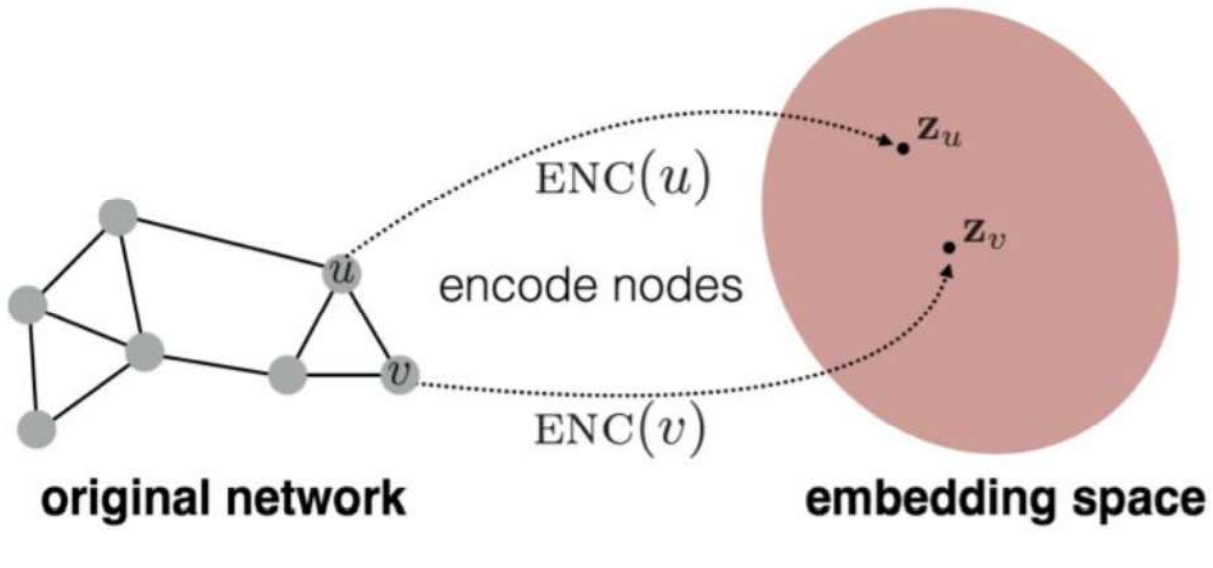

Figure 5: Embeddings of two nodes with respect to a proximity metric [6].

### 2.4.1a Matrix Factorization and Random Walks

A fundamental task in graph representation learning is neighborhood reconstruction; we begin with a brief overview of commonly used decoder models and loss functions. For an undirected homogenous graph with $|\mathcal{V}|$ nodes and a pre-defined node-node similarity matrix $\boldsymbol{S} \in \mathbb{R}^{|\mathcal{V}|\times|\mathcal{V}|}$, the goal is to learn embeddings $\boldsymbol{z}_u \in \mathbb{R}^d, \forall u \in \mathcal{V}$, such that we minimize the loss formalism:

$$\mathcal{L} = \sum_{(u,v)\in\mathcal{D}} \ell(Dec(\boldsymbol{z}_u, \boldsymbol{z}_v), \boldsymbol{S}[u,v])$$

Here, $Dec(\boldsymbol{z}_u, \boldsymbol{z}_v): \mathbb{R}^d \times \mathbb{R}^d \mapsto \mathbb{R}$ is a differentiable decoder which operates over pairs of node embeddings, $\ell: \mathbb{R} \times \mathbb{R} \mapsto \mathbb{R}$ is a specified loss function and $\mathcal{D}$ is a training dataset of node pairs. Over the years, there have been many instantiations of this model that seek to reconstruct the similarity matrix – in the simplest case defined as adjacency matrix $\boldsymbol{S} \triangleq \boldsymbol{A} \in \mathbb{Z}^{|\mathcal{V}|\times|\mathcal{V}|}$ – via matrix factorization [58, 59, 60, 61]. Alternatively, there are stochastic approaches that attempt to maximize the likelihood of random walk trajectories with respect to the learned node embeddings. We model the probability of visiting a node $v$ on a random walk of length $T$ from start node $u$ as:

$$Dec(\boldsymbol{z}_u, \boldsymbol{z}_v) = \frac{e^{\boldsymbol{z}_u \cdot \boldsymbol{z}_v}}{\sum_{w\in V} e^{\boldsymbol{z}_u \cdot \boldsymbol{z}_w}} \approx p_T(v|u)$$

Since the denominator is a costly $O(|\mathcal{V}|)$ to compute, DeepWalk [62] and node2vec [63] propose efficient approximations involving hierarchical softmax and noise contrastive sampling techniques, respectively. The embeddings are trained such that they minimize the cross-entropy loss over node pairs acquired from random walk trajectories, $\mathcal{D}$:

$$\mathcal{L} = \sum_{(u,v)\in\mathcal{D}} -\log Dec(\boldsymbol{z}_u, \boldsymbol{z}_v)$$

This can be also interpreted as maximum likelihood estimation of the observed trajectories in the latent space. Interestingly, random walk embeddings are theoretically connected to matrix factorization of non-linearly transformed symmetric normalized Laplacian matrix and is closely related to spectral clustering [64]. The strategies and concepts put forward by random walk methods offer a riveting outlook on planning applications; one could optimize the embeddings to reflect node statistics from task trajectories generated by classical planners. However, we recall that 3D scene graphs are multi-relational, and hence, they require additional graph learning techniques to process node features.

There are a variety of methods that have been proposed to decode multi-relational graph embeddings. A very simple extension to the aforementioned matrix factorization approaches is tensor factorization, where the objective is to minimize the reconstruction loss:

$$\mathcal{L} = \sum_{u \in \mathcal{V}} \sum_{v \in \mathcal{V}} \sum_{\tau \in \mathcal{R}} \| Dec(\boldsymbol{z}_u, \tau, \boldsymbol{z}_v) - \boldsymbol{A}[u, \tau, v] \|^2$$

In the above equation, $Dec(\boldsymbol{z}_u, \boldsymbol{z}_v): \mathbb{R}^d \times \mathcal{R} \times \mathbb{R}^d \mapsto \mathbb{R}$ is a multi-relational decoder, $\mathcal{R}$ is the set of relationship types, and $\boldsymbol{A} \in \mathbb{Z}^{|\mathcal{V}| \times |\mathcal{R}| \times |\mathcal{V}|}$ is the graph adjacency tensor. This loss function can be easily coupled with a selection of multi-relational decoders that are capable of representing symmetric relationships [67] anti-symmetric relationships [68], or both [69, 70, 71]. The pitfall of tensor factorization is that it incurs a $O(|\mathcal{V}|^2 |\mathcal{R}|)$ complexity and degrades the rather sparse edge classification problem to a dense binary regression. Thus, faster alternatives like cross-entropy loss with negative sampling [65] or contrastive estimation techniques [66] involving max-margin objectives are typically used in practice. The speed-up results from their reduced computational cost, which is associated with a cost much closer to $O(|\mathcal{E}|)$, where $\mathcal{E}$ is the set of graph edges.

### 2.4.1b Overview of Graph Neural Networks

The graph neural network encoder model enables us to extract informative node embeddings in a fully differentiable manner through iterative applications of the *Neural Message Passing* mechanism. In each round of neural message passing, the hidden embeddings of all nodes in the graph are updated in the generalized form:

$$\boldsymbol{h}_u^{k+1} = Upd^k\Big(\boldsymbol{h}_u^k, Agg^k(\{\boldsymbol{h}_v^k, \forall v \in \mathcal{N}(u)\})\Big), \forall u \in \mathcal{V}$$

Above, $Upd^k$ and $Agg^k$ correspond to the update and aggregation functions at message passing iteration $k$ [6]. This process updates the embedding of node $u$ by fusing its features with the features of its neighboring nodes $v \in \mathcal{N}(u)$. Repeating this essential GNN operation allows for the hidden features of each node to become increasingly representative of its $k$-hop neighborhood, much like the concept of increasing receptive fields in convolutional neural networks. The final node embeddings $\boldsymbol{z}_u, \forall u \in \mathcal{V}$ are simply assigned as the hidden embeddings $\boldsymbol{h}_u^T$ at the last message passing iteration $T$. The embeddings can then be used as desired, which could involve any of the previously mentioned reconstruction and relation prediction tasks, or alternative objectives for node, edge, or graph classification. Once losses are computed, they can be propagated throughout the network to update the learnable parameters. While the concepts of hidden embeddings and neural message passing are consistent amongst all GNN architectures, the main distinctions therein lie in the definitions of $Upd^k$ and $Agg^k$.

In its primitive form, the neural message passing mechanism can be expressed as a non-linear activation of the current hidden embedding summed with its local neighborhood features after they have been linearly transformed by weight matrices $\boldsymbol{W}_s, \boldsymbol{W}_n \in \mathbb{R}^{d_{k+1} \times d_k}$ [72]:

$$\boldsymbol{h}_u^{k+1} = \sigma\left(\boldsymbol{W}_s \boldsymbol{h}_u^k + \boldsymbol{W}_n \sum_{v \in \mathcal{N}(u)} \boldsymbol{h}_v^k + \boldsymbol{b}_u^k\right), \forall u \in \mathcal{V}$$

We can see that this equation closely resembles the layer-wise update formula commonly used in multi-layer perceptrons (MLPs). Equivalently, we can express the graph-level update for all node embeddings $\boldsymbol{H}^{k+1} \in \mathbb{R}^{|\mathcal{V}| \times d_{k+1}}$ by incorporating the adjacency matrix $\boldsymbol{A} \in \mathbb{R}^{|\mathcal{V}| \times |\mathcal{V}|}$ as follows:

$$\boldsymbol{H}^{k+1} = \sigma(\boldsymbol{H}^k (\boldsymbol{W}_s)^T + \boldsymbol{A}\boldsymbol{H}^k (\boldsymbol{W}_n)^T)$$

A particular concern with this model is that it yields embeddings that are heavily biased towards nodes with large neighborhood degrees. A common technique to mitigate the effects of varying node degrees is neighborhood normalization, which adaptively scales the aggregated neighborhood features such that their magnitude is consistent with respect to other neighborhoods in the graph. Neighborhood normalization assumes no extra parameters and has been used to stabilize the well-known Graph Convolutional Networks (GCNs) architecture [14]:

$$\boldsymbol{H}^{k+1} = \sigma\left(\boldsymbol{D}^{-1/2} (\boldsymbol{A} + \boldsymbol{I}) \boldsymbol{D}^{-1/2} \boldsymbol{H}^k (\boldsymbol{W})^T\right)$$

In particular, GCNs apply a symmetric normalization with the degree matrix $\boldsymbol{D} \in \mathbb{R}^{|\mathcal{V}| \times |\mathcal{V}|}$, in conjunction with self-loops allowing for weight sharing between nodes and their respective neighbors. This GCN framework is well known for its computational efficiency and is frequently used as a baseline in many graph representations learning tasks. Moreover, its architecture has been extended to multi-relational graphs, dubbed RGCNs [73], which uses edge-specific weights for neighborhood aggregation. Memory efficient variants of the RGCN model have been

proposed when $|\mathcal{R}|$ is large by imposing weight sharing with a linear combinations of learnable basis matrices:

$$\boldsymbol{h}_u^{k+1} = \sigma\left(\boldsymbol{W}_s\boldsymbol{h}_u^k + \boldsymbol{W}_n \sum_{\tau\in\mathcal{R}} \sum_{v\in\mathcal{N}_\tau(u)} \frac{\boldsymbol{W}_\tau \boldsymbol{h}_v^k}{g\big(\mathcal{N}(u),\mathcal{N}(v)\big)} + \boldsymbol{b}_u^{\boldsymbol{k}}\right), \forall u \in \mathcal{V}$$

$$\boldsymbol{W}_\tau = \sum_{i=1}^{|\mathcal{B}|} \gamma_{i,\tau}\boldsymbol{B}_i$$

Notice that when $\mathcal{B}$ is defined such that $|\mathcal{R}| \gg |\mathcal{B}|$, we can expect significant parameters savings with this model. Another interesting mechanism that can be incorporated in the design of GNNs is *attention* [74]. Graph attention networks (GATs) [75] embrace the idea of self-attention, and attempt to learn attention weights $\alpha_{u,v}$ to scale the local neighborhood embeddings as shown below.

$$\boldsymbol{h}_u^{k+1} = \sigma\left(\boldsymbol{W}_s\boldsymbol{h}_u^k + \boldsymbol{W}_n \sum_{v\in\mathcal{N}(u)} \alpha_{u,v}^k \boldsymbol{h}_v^k + \boldsymbol{b}_u^{\boldsymbol{k}}\right), \forall u \in \mathcal{V}$$

$$\alpha_{u,v}^k = \frac{\exp\Big(\boldsymbol{a}^T(\boldsymbol{W}_a\boldsymbol{h}_u^k \oplus \boldsymbol{W}_a\boldsymbol{h}_v^k)\Big)}{\sum_{w\in\mathcal{N}(u)} \exp\Big(\boldsymbol{a}^T\big(\boldsymbol{W}_a\boldsymbol{h}_u^k \oplus \boldsymbol{W}_a\boldsymbol{h}_w^k\big)\Big)}$$

While this is one of several ways to generate attention weights, the principal idea remains that node neighbors can be self-attended to as a function of their hidden embeddings. When analyzing scene graphs, the expressiveness of GATs may work to emphasize relationships between objects and places in *task space*, when they may not necessarily be heavily related in *metric space*. Correlations amongst the learned attention-driven embeddings could be used to draw powerful search heuristics for planning at relatively low-cost in comparison to hand-crafted heuristic functions. This type of graph processing could be made possible by multi-relational attention models, recently proposed by *Teru et al.* [76]. Its important to note that the aggregation functions employed in GCNs, RGCNs, GATs are specific instances of permutation invariant set aggregators [64, 77], and that MLPs can technically be employed as a universal set approximator [78]. In addition, researchers have also studied the benefits of permutation sensitive set aggregators like Janossy pooling [79].

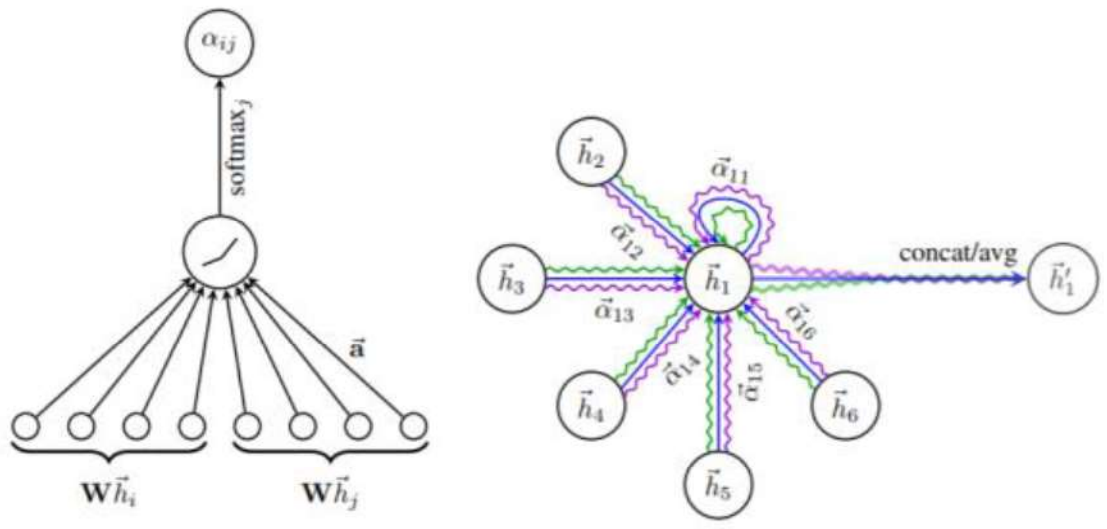

Figure 6: Neural message passing in graph attention networks [75].

As with convolutional neural networks, many similar cross-layer feature aggregation methods have been applied to GNNs showing improvements in terms of gradient flow, expressiveness, and prevention of over-smoothing, i.e., the uniformity of node embeddings in deep GNNs. Approaches include, but are not limited to: skip connections and feature concatenation [57], linear interpolation [80], gated recurrent updates [81, 82], and jumping knowledge connections [83]. Additionally, graph classification and regression tasks require a global descriptor that can computed from the learned embeddings through averaging. Sophisticated approaches make use of LSTMs to iteratively predict attention weights, which are in turn used to aggregate the node embeddings into a global feature [84]. For more on the real-world applications of GNNs along with common knowledge tips for training and implementation, we refer readers to [6, pp. 68-74].

### 2.4.1c Case Study: Object-centric Reasoning with GNNs

This section provides an overview of three prominent applications of Graph Neural Networks to object-centric relational reasoning tasks [85, 86, 87]. These methods ground the expressive power of GNNs in concrete examples that require inductive and abstract reasoning to infer relationships and continuous properties of objects in realistic scenes and dynamical systems. In this context, they hold a particular relevance to learning object-centric representations of 3D scene graphs, although, their designed use cases are not directed towards robot task planning.

We begin with Neural Relational Inference (NRI) [85], an approach to learning the underlying interaction graph between objects in a dynamical system in a fully unsupervised fashion. The method involves two GNNs: an encoder model that predicts that latent interaction graph describing the relationships between objects from a densely connected input graph, and a decoder model which iterates on neural message passing through the latent graph structure to infer the continuous states of objects in future timesteps. These models are casted into a variational graph autoencoder (VAE) framework [88], as shown in Figure 7. The objective is to maximize the reconstruction likelihood of future states of objects and minimize the Kullback-Leibler Divergence between the approximate conditional distribution of edges given the input graph, and a uniform prior over the edges. The variational approach ensures that the inferred interactions graphs of similar inputs will be sufficiently close in the embedding space, while the interaction graphs of dissimilar inputs will be sufficiently diverse. Prior this work, methods attempted to predict system dynamics in end-to-end fashion by assuming that the input graph is fully connected, and learning an *implicit* structure of the interaction graph. By partitioning this setup into a two-step process, whereby we derive an *explicit* interaction graph before predicting the system dynamics, NRI manages to significantly outperform its counterparts.

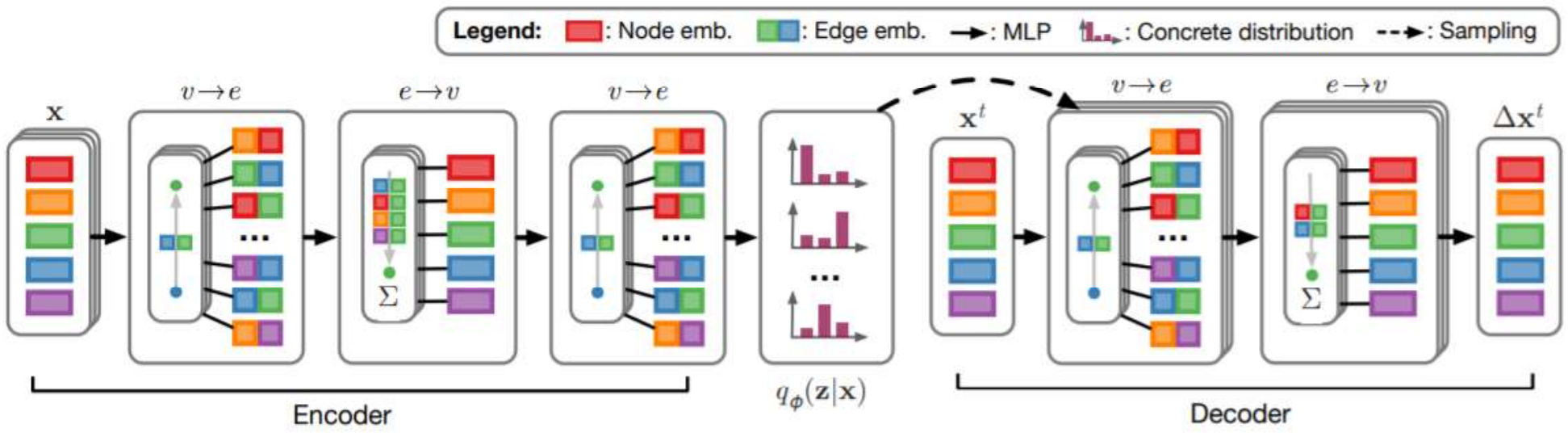

Figure 7: NRIs' graph VAE model for learning explicit interaction graphs [85].

The authors demonstrate the utility of NRI on small-scale problems such as learning the interaction graph and predicting the future states of a spring system, and charged particles in a constrained environment. In the planning regime, if we interpret the support structure of 3D scene graphs as a dynamical system with logical transitions defined by the actions of a policy, it would be very interesting to observe the latent interaction graph predicted by NRI. In the ideal case, if the input graph is conditioned on a task type and the goal condition, we might expect the inferred interaction graph to describe the set of logical predicates most relevant to the task, or an abstract form of object-centric task affordances. In spirit, this mimics the graph sparsification and attention objectives of CAMPs [12] and PLOI [7]. However, they supervise the training of their predictive models offline with variable or object labels as required by an expert planner to solve a task. NRI would instead use the online changes of the scene graph throughout an expert demonstration to learn a meaningful latent structure. Although, there are some uncertainties with this approach, as NRI would be operating on graphs that are two orders of magnitude larger than those presented in the paper, and it may be computationally infeasible to model $k^2$ relationships in the encoder. Furthermore, the scene graph transitions induced by object-parameterized actions are very sparse over a few timesteps, and one would likely need to model state transitions over longer horizons. From a planning perspective, this is desirable, as each pass of the NRI model could deduce the set of logical facts required to reach the next subgoal in online fashion, which is far more flexible than predicting the set of sufficient objects or ground-actions at the start of task.

Recent works by *Kipf et al.* [86] and *Locatello et al.* [87] bridge us into self-supervised learning of abstract state representations directly from pixels. Similar to the NRI setup, the objective in this class of methods remains to predict the action dependent future state in a trajectory. Slot Attention [86] is a differentiable interface between perception modules and feature containers called slots. The concept of using slots to represent object-centric features provides a strong inductive bias for tasks that require object or group-level reasoning, in comparison to modelling interactions directly from pixel-space features. The module was designed to be simply added on top of a CNN feature extractor, and iteratively applies a modified scaled dot-product attention [74] to aggregate $N$ image space feature vectors into $K$ slots, which can then be used for a downstream task. This work demonstrates its use for unsupervised object discovery under the reconstructive objective of pixel-space dynamics modelling. In this scenario, each slot is used to reconstruct an exclusive region of the future state, which forces the slot attention module to extract localized representations of objects in the latent space. While the paper makes no implication on the use of GNNs in their model, we note that the scaled dot-product attention mechanism is technically a GNN. Here, one could interpret the localized CNN features as nodes of a graph, and the query-key-value feature aggregation step as attention-based neural message passing in a fully connected graph.

Slot attention, like many other image space reconstruction techniques, suffers drawbacks on the precision of their predictions. Since the commonly used image space loss functions place equal weight over all pixel residuals, the cost of neglecting a small but potentially important part of state is almost negligible. Moreover, in real-world applications, the image formation process is highly sensitive to parameter variations or domain shifts. As a result, pixel-space models typically produce coarse reconstructions that maximize the likelihood of large, static, and easy to capture regions of the image. This observation motivated the design of Contrastively-trained

Structured World Models (C-SWMs) [87]. Instead of state reconstruction in the image space, C-SWM formalizes a noise contrastive loss in a latent state space, and a GNN instantiates the latent transition model for each object in the scene. Hence, the state of the system is actually a factored state of object-centric features extracted by a CNN; relating to scene graphs in which case the overall state is semi-factored to account for hand-crafted relationships between objects. The drawback of C-SWM is that the proposed architecture limits the number of distinct objects that can be modelled, and no explicit inference on the interaction graph is made, which may make it challenging to scale up to larger physical systems. Although, the idea of contrastive representation learning at the level of objects is a promising direction, and is one that associates well with our pursuit of learning scene graph representations for more capable planning.

### 2.4.2 Learning to Plan in Structured World Models

In the analysis of the scene graph structure, it is clear that they induce metric proximity. Objects are grouped into hierarchies according to shared spatial and physical relationships. However, this grouping does not necessarily reflect task proximity. We can think of many examples of object pairs, that when conditioned on a task, become related in terms of their affordance. For example, the relationship between clothing and a laundry machine only becomes apparent when the clothing is dirty and needs to be washed. Therefore, while the claim that relates the local arrangement of entities to their task space proximity may hold for some objects (e.g., tea bag, kettle, cup), the general principle is not sound. Furthermore, it is the local density of related objects that contributes to an excessive branch factor, deterring the efficiency of classical planners. To this end, attending to certain regions of the 3D scene graph predicated on a task and goal condition has the potential to substantially improve the speed of planning, especially if one can manage to abstract only the relevant parts of the factorized state space.

#### 2.4.2a Learning Projective Abstractions for Planning

To build on this notion of graph abstraction, we turn to the concept of sparsification, which is a well-studied domain in classical graph theory that investigates sparse sub-graph approximations to dense graphs [89, 90]. Only in this case, we are not interested in the sub-graph approximations for the sake of reconstruction, but rather, for how effective they are in terms of describing a simplified planning problem. Underpinning this motivation is an assumption on the class of planning problems that we would like to approach; that is, the planning problem must have a state space that resembles a graph structure. Fortunately, there are many domains that range in both size and complexity which fit this narrative [30].

Two pieces of literature that explores this very idea are given by *Silver and Chitnis et al*. [7, 12], under the aliases of contextual irrelevance and object importance. Starting with [12], we build an intuition on what it means for variables to be irrelevant. A factored Markov Decision Process (MDP) maintains many of the familiar components of a standard MDP: a finite set of states and actions, a state-action parameterized transition model, a reward function and a finite horizon. However, it benefits from the added property that each state and action can be factored into a finite set of variables [91], and thus, a state is an assignment of values over all of its variables, much like the definition of a state in the planning problem outlined in Section 2.1.1. However, unlike the objectives of a classical planning problem or a standard MDP, the objective here is to find a policy that minimizes the expected discounted future rewards minus a compute-cost metric which accounts for the wall-clock time of evaluating the policy.

$$J(\pi, w) = \mathbb{E}\left[\sum_t R(s_t) - \lambda \cdot ComputeCost(\pi, s_t)\right]$$

This objective function introduces a clear trade off between the optimality of the solution and the efficiency of the policy. A pure policy learning approach which incurs very little compute time will often struggle to achieve high rewards for long-horizon tasks. Conversely, a pure planning approach which is able to trace the effect of its actions far into the future to achieve high reward solutions is hindered by this timely procedure. Thus, in this setting, the optimal policy likely lies somewhere in between the above two approaches, as depicted in Figure 8.

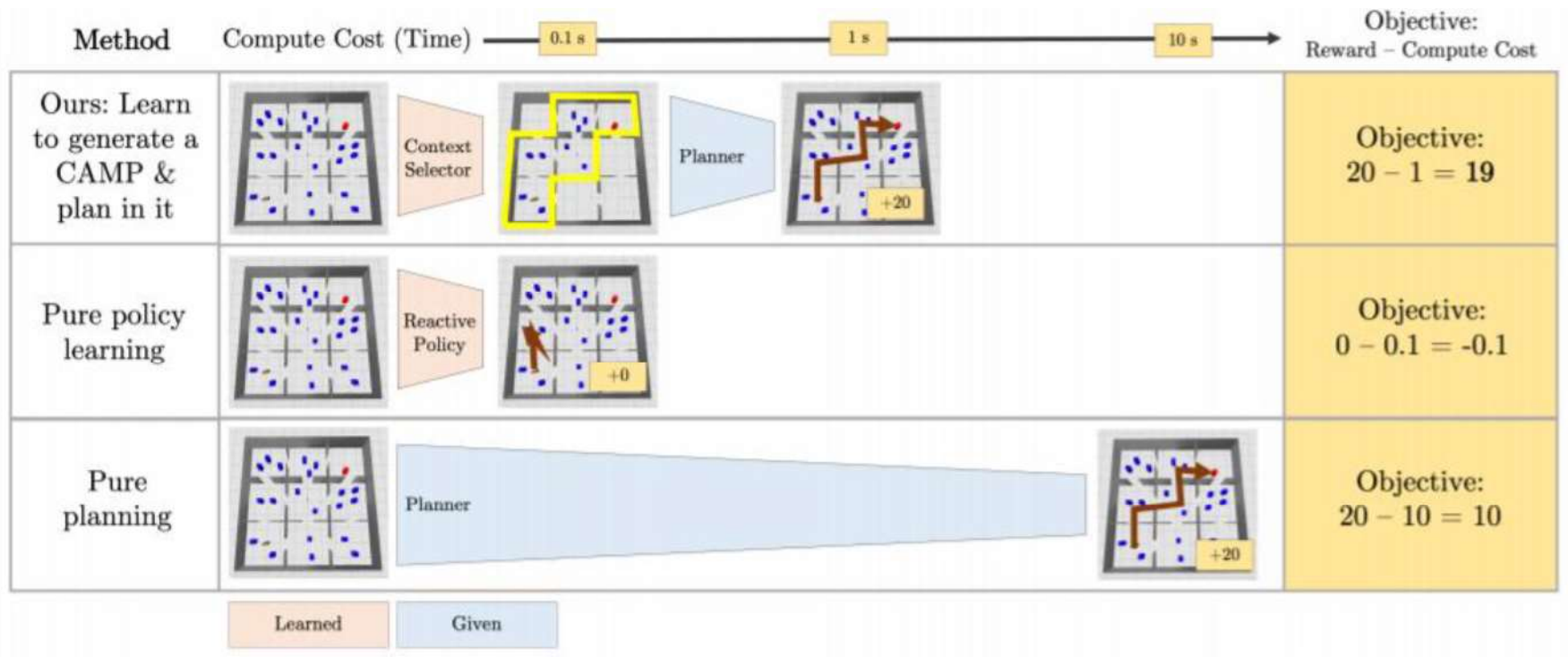

Figure 8: CAMPs' approach to solving factored MDPs [12].

Now, an *irrelevant* variable is a term which defines state or action variables that can be removed from the problem without having any affect on the possible rewards that can be obtained by a policy given a task. Unfortunately, the set of irrelevant variables is usually quite small, as one can imagine many scenarios in which a policy could act such that a variable initially thought to be redundant for a task becomes relevant. Thus, *Silver and Chitnis*. [12] generalizes the idea of *irrelevance* with *context-specific independence* (CSI) [92]. Together, they define variables that are *contextually irrelevant*. For instance, the expected rewards of completing a kitchen task are unaffected by state and action variables of the backyard if and only if the robot is confined to the kitchen (context). Projecting out all contextually irrelevant variables transforms a factored MDP into an abstract problem, which they refer to as a Context-specific Abstract MDP (CAMPs). We note that projective abstractions have been used in other works [93, 94, 95], and that there also exists less crude abstraction techniques [96, 97] compared to simply dropping irrelevant variables – although, they have not been explored in learning-based planning.

CAMPs are not guaranteed to contain the optimal solution of the original MDP (in fact, completeness and soundness are not guaranteed either), but the hypothesis is that planning over the reduced state and action spaces yields computational benefits which affords less optimal solutions. The order of operations is then: (1) identify a suitable context for a task, i.e., a constraint on a subset of variables in the factored MDP, (2) project all contextually irrelevant variables out of the problem to obtain the CAMP, and (3) execute the planner. Contextual independence is approximated with a brute-force sampling procedure which empirically tests whether one sampled set of variables in the current timestep affects another set of variables in a future timestep. Likewise, contexts are generated through another sampling procedure which builds a pool of conjunctions and disjunctions of variables up to a certain length. Conditioned on the task, we would like the robot to self-impose a context that improves planning efficiency. They train an MLP for multi-class context classification, where context labels maximize the reward-compute time objective and are generated offline. The reported results indicate that planning over context-specific abstract MDPs yields high reward solutions at low computational cost, and outperforms competing baselines on numerous challenging robotic TAMP domains.

As the authors point out, this hybrid method has several drawbacks. The context selector is limited to classifying over a finite set of discrete logical constraints, when in fact, many of the useful constraints in robot applications are over continuous variables. In addition, this method only generates one constraint per planning problem, and offers no recourse if planning fails over the abstract MDP. Only weak generalization is tested; a validation or test problem is defined by a rearrangement of the initial state and goal states in a training sample, not a different task entirely. Most importantly, using an MLP on the featurized state for context selection fails to capture the inherit relational structure of object-centric planning domains, which could serve as an inductive bias on the learned representations and promote stronger generalization. Herein lies the contributions of PLOI [7], which addresses these concerns in the domain of PDDL task planning.

The factored MDP formalism does not provide any relational information between objects to simplify the deduction of their state variables. Instead, the context selector, a feed-forward MLP, is trained to classify one of many possible constraints on the variables from a featurized state of the environment, such as flattened vector of image pixels. This setup limiting, in that it forces the validation and test problems to be sufficiently similar to the training instances. In PLOI [7], a planning problem is modelled in terms of the set of objects it contains and the relationships between them in a predicate logic-based graph. The notion of contextual irrelevance at the level of individual variables is evolved into the concept *importance scores* at the level of objects. In the process, the challenge of creating useful projective abstractions for planning, which was initially a one-of-$K$ context classification problem (where $K$ is large), has been transformed into a binary node classification problem of an object-centric graph. Hence, PLOI proposes a graph neural network architecture that takes as input a graph of objects connected by their initial state relationships and the task (implicitly embedded with the goal configuration), and predicts a one-dimensional embedding for each object. After a sigmoid activation, the node embeddings represent the importance of each object for a particular task. Thresholding the importance scores to acquire a simplified planning problem is equivalent to: (a) projecting out all state variables associated with that object in CAMPs; (b) pruning all possible ground-actions on that object, which immensely reduces the branch factor of the planning graph. Planning can then be carried out in the abstract problem by a heuristic planner of choice.

A sample of the PLOI method in the Blocks domain is illustrated in Figure 9. As shown, the relational structure of objects is parameterized in terms of their properties and predicates, which in most classical planning domains corresponds to their semantic class and support structure, e.g., *block(A)* and *on(A, B)*, respectively. Notice that this model can be easily extended to real-world environments represented by 3D scene graphs, as continuous properties of objects such as their poses and volumes can be appended as node features. Furthermore, continuous pairwise relationships can also be factored into this model as additional edge features. As we know, graph neural networks can be applied to input graphs of arbitrary size, so long as the node and edge features are consistent with the expected input dimensions. Since this is the case in each planning domain (feature dimensions are domain-specific), then the GNN architecture can technically be applied to problems of arbitrary sizes as well – overcoming a fundamental limitation of the CAMPs model. Analysis of the PLOI results in deterministic planning domains indicate that inductive biases learned by the GNN model from smaller problems are able to generalize to problems consisting of many more objects, relations, and more complex goal configurations. However, the compositional structure of each domain is mainly independent of the problem size, which may explain the generalization characteristics of their model. Further conclusions on generalizability would require an extended study of the model's performance on out-of-distribution samples.

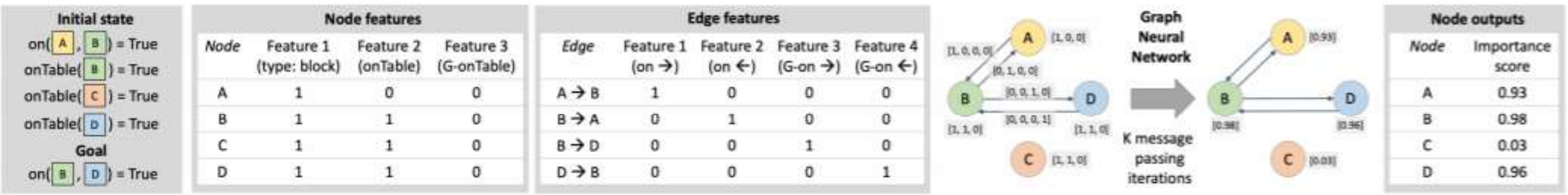

Figure 9: Logic-based graph utilized by PLOI's GNN for importance scoring [7].

The model is weakly supervised by sufficient object sets. A sufficient object set is defined as the reduced subset of objects such that a valid solution still exists when planning with just those objects. While finding the minimal object set for each training instance is infeasible in reasonable time, the authors propose a greedy search strategy: remove an object at random, attempt to plan with the reduced object subset, repeat the procedure if a valid plan is found, otherwise return the previous object subset. Because the planning time of traditional heuristic planners is exponential in the number of extraneous objects, removing a conservative number of irrelevant objects can make a significant impact on planning efficiency. Hence, the GNN is trained through a weighted binary cross entropy loss as follow:

$$\mathcal{L}_{PLOI} = -\sum_{\forall n \in \mathcal{V}} \lambda p_n \log[\sigma(z_n)] + (1-\lambda)(1-p_n)\log[1-\sigma(z_n)]$$

Here, $z_n$ is the predicted importance logit for object $n \in \mathcal{V}$, and $p_n$ is the target acquired from the sufficient object set. During inference, a hyperparameter, $\gamma$, is used to threshold objects based on their predicted importance. This offers a trivial route for recourse should planning over the one-shot object set fail; simply decrease the threshold, $\gamma_{k+1} = \gamma_k \cdot \gamma$, append the newly satisfying objects to the subset, and re-attempt planning. So, the PLOI method is complete in the limit. Although, each failed planning attempt incurs an undesired cost, and it would be preferable to obtain a plannable object set after the first threshold. We may expect the precision of predicted object importance scores to improve with the quality of supervision.

In summary, the PLOI model exemplifies the utility of a hard-attention mechanism at the object-level which can be trained offline and applied prior to executing a classical planning algorithm. The method leverages a GNN node classifier, trained on sufficient object sets, to factor in logical relationships between objects when making projective abstractions. However, the approach is not without its disadvantageous, a primary one being its inability to replan in dynamic and stochastic environments which is evident in their results. In addition, each problem domain requires a new model to be trained, while having a single model to facilitate faster planning across a variety of tasks would be advantageous. Sub-optimal supervision makes it difficult to effectively threshold importance scores, and no attempt at modelling prediction uncertainty is made. Overall, the empirical results are impressive, particularly in deterministic domains, proving to be an appropriate baseline for our research. In addition, they provide helpful comparisons to action grounding methods via Inductive Logic Programming [10] and modified GNN architectures that predict ground action scores as an alternative to object importance scores.

#### 2.4.2b Neural Symbolic Planning

The approaches covered in the previous section [7, 12] focus on abstracting the important parts of a domain prior planning with classical algorithms. In this section, we will discuss a different branch of learning-to-plan methods that rely solely on deep neural networks to predict state transitions or to learn value functions over states in an online search. Oftentimes, these planning systems are modelled after successful search strategies in classical planning (i.e., regression planning, Monte-Carlo Tree Search) by replacing key components, like the heuristic evaluation function, with neural network modules which leverage experience during inference. Approaches of this class have already shown their merit, from becoming the world's top performer in challenging games like Go [98, 99], to mastering virtually any game through self-play [100]. Yet, these approaches have only recently made their way into robotics applications, which constitutes its own set of challenges – continuous state and actions spaces, partial observability, and environmental stochasticity and dynamicity. Thus, the preliminary works discussed below are limited to learning in structured domains or constrain the task to require only a discrete set of logical actions while neglecting physical interactions and low-level motion.

Model-based approaches to robot planning attempt to learn image or latent space dynamics models which are used to simulate the effect of action sequences forward in time. The success of these methods is deterred by the accumulation of error in the dynamics models' predictions over long-horizons, which makes it impossible to construct reliable plans in different task settings and environments. Moreover, converting a goal definition in the form of an instruction, logical expression, or configuration into an observation which can be used to condition the policy is non-trivial. Regression planning networks (RPNs, *Xu et al.* [13]) offers a new outlook to online planning in object-centric structured domains, like 3D scene graphs, without requiring explicit action operators, planning domain definition, or symbolic states. They adopt a regression planning strategy, which attempts to plan backwards from the goal instead of planning forwards from the initial state. In doing so, the current observation can be used to condition the backwards search in symbolic space. Full access to the initial state observation supports more accurate prediction and ordering of subgoals, which are handled by the precondition network and subgoal serialization algorithm depicted in Figure 10. Subgoals are prioritized according to relation predictions from a dependency network, which forms a direct acyclic graph (DAG) of dependencies between logical facts (subgoals) of a state. A topological sort of the DAG is then used to return the next subgoal to be expanded. Since each intermediate goal state in the backwards search is a partial state under the closed world assumption, a reachability network is used to classify whether the current goal state can be reached from the initial state with a single motion primitive. Their robot task planning experiments demonstrates RPNs' impressive ability to solve compositional tasks in small-scale scene graphs, although the method's applicability to large scenes is yet to be tested.

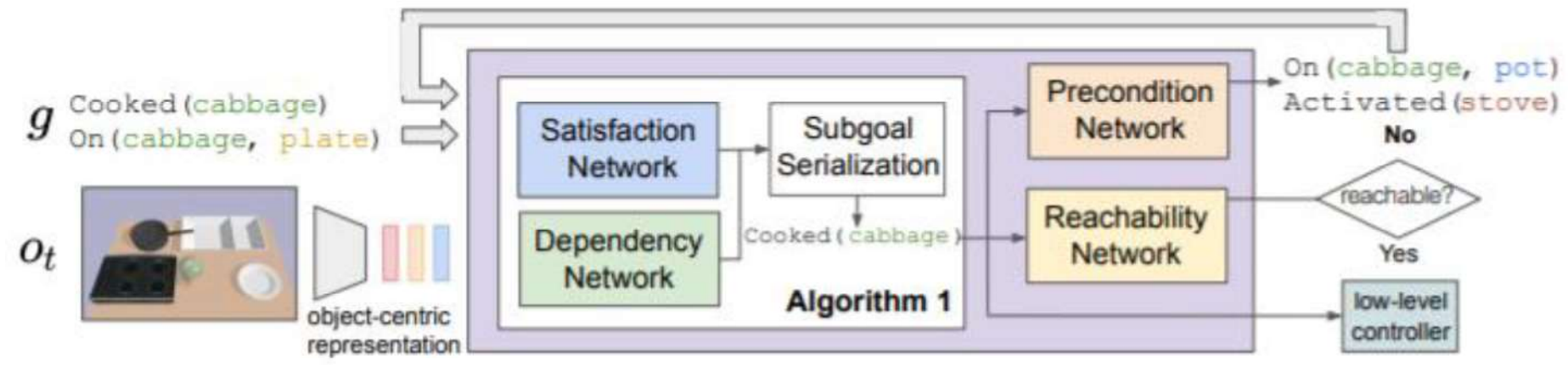

Figure 10: Regression planning network architecture for scene graph planning [13].

On the other hand, methods like MCTSNets (*Guez and Weber et al.* [23]) have had remarkable success in long-horizon planning, albeit outside the field of robotics. However, bridging statistical tree search methods with function approximation holds promise for task planning in 3D scene graphs, which simply extends a predicate-based relational structure with hand-selected unary and pairwise properties. In this approach, all original components of MCTS are replaced with three fully differentiable modules for state evaluation, simulation policy, and value propagation. Node-level statistics of the tree are implicitly represented by embeddings, and are updated by the rewards and child node embeddings at each simulated iteration. As such, the search procedure differs drastically from Value-Network MCTS, which simply replaces the need for simulation with a value regression network and uses the same upper confidence bound (UCB1) formula to sample actions in future iterations. In contrast, MCTSNets is autoregressive, and the tree expansion corresponds to a stochastic feed-forward neural network (i.e., sampled actions), where the computation flow is determined by the final tree structure. These modules are optimized in supervised setup through a dynamic computation graph, where stochastic actions sampled during simulation are reinforced by the log-likelihood of ground-truth actions acquired from expert demonstrations. Such an approach is applicable to scene graphs given expert demonstrations from classical PDDL planners. One issue, however, is the dependency on dense rewards which is not currently attainable in a simulator that supports robot task planning in scene graphs. Hence, to encourage experimentation with these online planning techniques, the design of rewards across various task types in a shared environment should be considered in future benchmarks.

Few works have extended these concepts to task planning in 3D scene graphs. *Zhu et al.* [17] leveraged RPN in their end-to-end hierarchical TAMP framework, but test cases only contained 2-6 objects. Moreover, the majority portion of their work focused on symbol mapping from RGB images to a symbolic scene graph for the RPN, and symbol grounding from symbolic space to a 3D geometric graph for motion planning. Lying in between PLOI and the objective of our research, Hierarchical Mechanical Search (HMS, *Kurenkov et al.* [16]) presents a method for object search and retrieval in partially observable scene graphs. In their framework, the scene graph is comprised of only two types of entities: containers that can be searched, and objects that can be moved. Initially, the scene graph is comprised only of containers (e.g., rooms and receptacles), but object node attributes are appended as leaf nodes upon searching a container. The goal is to identify a path from the root node of the scene graph to a leaf node that is the target object. This is achieved with a greedy depth first search strategy guided by a node-level attention heuristic. There are several parallels to PLOI, namely: (a) the node-level heuristic is inferred from the relational structure of the domain; (b) a threshold is set to project-out irrelevant parts of the scene graph; (c) the threshold is decreased to incorporate more of the original graph should planning fail. Unlike PLOI, the proposed neural message passing mechanism operates on one level of the hierarchy at a time. Figure 11 illustrates the mechanism for aggregating the embeddings of a parent node and its child nodes to predict a binary importance score. Notice that their depth first search method exploits the scene graph's hierarchical structure. Such a search strategy is not applicable to our tasks where action transitions equate to modifying attributes of the scene graph, rather than traversing its nodes (i.e., scene graph versus planning graph, as described in Section 2.1).

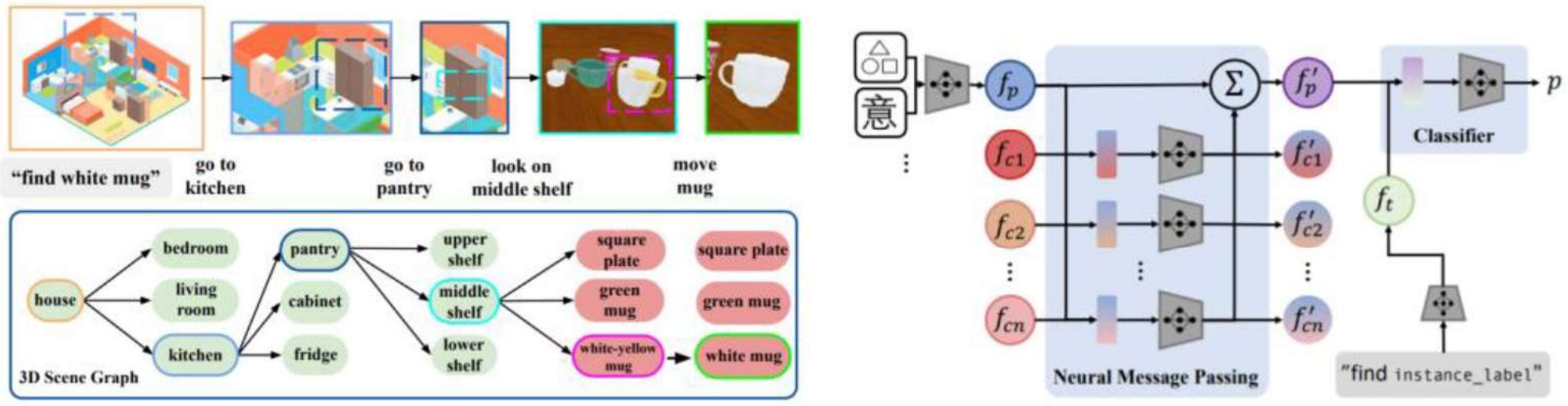

Figure 11: Object retrieval in scene graphs (left) with hierarchical neural message passing (right) [16].

Despite the clear distinction between the object retrieval task of HMS and our scene graph rearrangement task, there are multiple techniques proposed that could potentially be synergetic to long-horizon hierarchical planning. The first is the level-order application of neural message passing, which resembles abstract or top-down reasoning akin to how humans might attempt to deduce the necessary rooms before receptacles to visit, and receptacles before objects to interact with. This procedure also better exploits the hierarchy in 3D scene graphs, giving the potential to save on computation by pruning nodes at higher-levels of the hierarchy, which in turn prunes all subsequent child nodes before having to consider their interconnections. Algorithmically, such a process would combine the strengths of HMS and PLOI to reduce the state space for planners. Furthermore, HMS exemplified how GNNs could be used on dynamically growing scene graphs, which could likewise extend planning support to partially observable environments. In this case, new scene graph representations could be generated for replanning upon discoveries of new rooms, receptacles, and objects.

The evident research gap in the use of learned graph representations to simplify long-horizon task planning in large scene graphs is one that will differentiate our approach from the rest. Specifically, [13, 16, 17, 23] supervise the training of online search (challenging to scale) with expert demonstrations, and the training of networks used in [7, 10] is dependent on an intermediate and sub-optimal supervision signal (sufficient object sets) and occurs in isolation from downstream planners. A subtlety between these two general approaches is to learn task-conditioned scene graph representations in a coupled optimization process that promotes stronger synergies between the graph neural network model and the downstream planner.

# Section 3: Methods

This section presents a summary of our methodologies, which can be categorized according two objectives. Our first objective is to identify and test a suitable robot task planning benchmark and justify its selection amongst the release of many high-fidelity simulators for the 3D scene graph planning task. In parallel, towards the objective of improved graph learning techniques for fast and robust planning, we perform a sensitivity analysis on PLOI [7], conduct further experiments with a modified GNN architecture, and suggest improvements to the overall framework including the input graph conditioning and the training scheme. As a first step in the process, we limit the experiments to the Blocks domain in PDDLGym [30]. However, our design decisions will serve as a foundation for future extensions to scene graph style domains.

## 3.1 SGPlan: A Scene Graph Task Planning Benchmark

### 3.1.1 Environment Overview

In pursuit of developing a task planning benchmark that encompasses a broad range of tasks and provides adequate support for both classical and learning-based approaches, there are several factors to consider. An important requirement, as to align with the underlying objective of this research, is that the domain should encourage the use of 3D scene graph structures and features to benefit task planning, rather than limiting the policy to ego-centric visual observations and or language directives. This implies that the tasks and scenes in the environment should be large and diverse enough to foster a need for graph-level reasoning, and perhaps, graph representation learning to efficiently find solutions. Hence, we would require scenes to contain on the order of a hundred objects at minimum, in an environment that enables interaction between the agent and objects of various class categories. Moreover, it would be preferable for each scene to have multiple rooms, and for rooms to be partitionable according to spaces or accessible receptacles, allowing for actionable abstractions in the form of 3D scene graphs. As described in Section 2.2, access to a number of object and scene level properties are a necessity for constructing 3D scene graphs akin to [4, 5]. If such a domain were to present a challenge to state-of-the-art task planners, an important subsystem of embodied AI agents, we could expect to ignite research at the intersection of the scene graph and planning communities. As such, we must first identify a planning environment that best meets the outlined high-level requirements. Table 1 summarizes the distinguishing features of candidate hosts for the scene graph task planning benchmark.

Table 1: An overview of environments for embodied AI and robotics.

| | *AI2-THOR [31]* | *Gibson [101]* | *iGibson [102]* | *AI Habitat [33]* | *Sapien [34]* | *VirtualHome [44]* | *TDW [103, 104]* | *VRR [105]* |
|---|---|---|---|---|---|---|---|---|
| *Scale* | Room (AC) | Building (RW) | Building (RW) | House (RW) | Room (RW) | Apartment (AC) | House (AC) | Room (AC) |
| *Scenes* | 120 | 1400 | 15 | - | - | 49 | 15 | 120 |
| *Objects* | 609 (DM) | - | 570 (DM) | - | 2346 | 308 | 200 (DM) | 72 (DM) |
| *Interactions* | F, PA | None | F | None | F | F, PA | F | F, PA |
| *Simulation* | RBP, PE | RBP | RBP | RBP | RBP | PE | RBP, PF | RBP, PE |
| *Integrated MP* | No | No | Yes | No | No | No | No | No |
| *Speciality* | TP, OS | Nav | PI | Nav | TAMP, AO, RT | TP, OS | TAMP, A, F | TP, OS |
| *Support* | Active | Moderate | Active | Active | Active | Inactive | Active | Active |

**Scale:** Artificially created environment (AC), and real-world environment (RW)
**Objects:** Annotated with dynamical and material properties (DM)
**Interactions:** Contact forces (F), predefined macro-actions (PA)
**Simulation:** Rigid-body physics (RBP), preconditions and effects (PE), particle simulation and fluids (PF)
**Speciality:** Task planning (TP), Task and motion planning (TAMP), physical interaction (PI), articulated objects (AO), object states (OS), navigation (Nav), ray-tracing (RT), audio (A), fluids (F)

In order to make the benchmark *attractive* to the vision and planning communities, it must provide strong support for 3D scene graphs such that they could be easily constructed and exploited for task planning. The tasks should be described in a language or format conducive with the common practices of the planning community, and the task evaluation protocols should align with those proposed in Rearrangement [18]. For the benchmark to be *useful* to the community, the tasks and scenes it presents should come from realistic and diverse distributions at a scale equal to or exceeding that of the existing benchmarks.

Many new simulators emphasize physical interaction between robots and objects [34, 102, 104], and several others tailor to robot navigation in environments created from the real-world [33, 101]. However, these environments call for the development of improved task and motion planners or low-level action policies to plan a path through the configuration space in order to achieve a task. Despite the need for some degree of task planning or hierarchical planning in these domains, the extent to which these modules are tested is limited by the motion planning and modelling of physical interactions aspects of the supported problems. The later subsystems, particularly in the learning regime, are far from the maturity of classical planning methods. Hence, the term "long-horizon" conferred on tasks in these benchmarks are accurate with respect to the entire robot system, but the tasks are hardly long-horizon with respect to the high-level planner. As a result, these domains typically offer very weak PDDL support, if not non-existent, and manually appending problem and domain files to create a parallel set of planning tasks would not present much of challenge for classical planners [20, 21].

There are exceptions, however, which include the AI2-THOR [31] and VirtualHome [44] simulators. As shown in Table 1, these domains remove the dependency on low-level actuation and introduce pre-defined macro actions that are simulated via preconditions and effects, much like the PDDL planning protocols. Moreover, the ability to alter object states through macro actions makes for a more interesting and challenging set of tasks (e.g., non-reversable state transitions). In fact, both offer a large and diverse set of tasks that could be easily casted into the Rearrangement framework. Unfortunately, they also share the same drawback in that they are comprised of room or apartment scale scenes, which does not compare in terms of scale to Gibson and its extensions [101, 102]. Thus, composing hierarchical scene graphs in AI2-THOR and VirtualHome yields rather flat representations, as they miss the *house* and *room* levels of abstraction that would facilitate large-scale rearrangement tasks. This aside, they both offer the fundamental buildings blocks for a scene graph task planning benchmark. We select AI2-THOR as the underlying simulator for our task planning experiments, given that it provides objects with dynamical and material properties, consists of more scenes, and is actively maintained.

There are multiple environments built on AI2-THOR which are aimed at certain niches of the embodied AI research sphere, and hence, the tasks they defined and the features they offer can vary quite significantly. Most recently, the Visual Room Rearrangement [105] benchmark was released for rearrangement type tasks. However, they offer no PDDL support and the challenge itself is geared more towards the full POMDP task definition of Rearrangement, while we are only concerned with the high-level planning component. In contrast, ALFRED [11] was developed for learning action policies from natural language instructions. Thus, they provide expert demonstrations for all tasks on their benchmark. The expert demonstrations are acquired by sampling tasks in the AI2-THOR environment, casting them into PDDL format, and executing a classical planner on them. ALFWorld [106] is a recent extension that aligns the ALFRED simulator state with abstract textual representations produced by TextWorld [107]. Their planning benchmark shares identical task categories to ALFRED, but at a much larger scale. In addition, to make the task generation process more efficient, their PDDL domains have received several updates which further abstract motion (i.e., 2D grid search) from the task planning problem. In this case, point-to-point demonstrations for robot motion between receptacle locations are generated by an A* planner after the task planner has determined which objects and receptacles to interact with. However, learning-based agents on the official benchmark are still obliged to produce grid-level motion actions. This distinction in the action space for the task planner is desirable, as it more strongly corresponds to the nature of task planning in 3D scene graphs, where the agent's position in symbolic space is confined to pre-defined locations of the graph (i.e., receptacles, places, rooms, buildings). Therefore, with full PDDLv2.2 [53] support, a scene graph suited action space, and a hierarchical state space, ALFWorld [106] will be used to benchmark classical planners for rearrangement in the room-sized scenes of AI2-THOR [31]. From this point on, we refer to the benchmark as the Scene Graph Planning (SGPlan) environment.

Table 2: Task distribution of the SGPlan benchmark.

| | *LoiL* | *PaP* | *SaP* | *P2P* | *PCoP* | *PHeP* | *PClP* | *Total (split)* |
|---|---|---|---|---|---|---|---|---|
| *Train* | 692 | 988 | 988 | 1063 | 892 | 893 | 858 | 6374 |
| *Validation* | 19 | 46 | 0 | 45 | 28 | 25 | 37 | 200 |
| *Test Seen* | 29 | 46 | 34 | 33 | 38 | 34 | 37 | 251 |
| *Test Unseen* | 54 | 30 | 33 | 24 | 36 | 42 | 36 | 255 |
| *Total (task)* | 794 | 1110 | 1055 | 1165 | 984 | 994 | 968 | 7080 |

**Tasks:** Look at object in light (LoiL), Pick and place (PaP), Stack and place (SaP), Pick two and place (P2P), Pick cool and place (PCoP), Pick heat and place (PHeP), Pick clean and place (PClP)

SGPlan contains 120 different scenes, as in AI2-THOR, which is composed of 30 distinct kitchens, bedrooms, bathrooms, and living rooms. The scenes contain on the order of a hundred objects, and object locations on receptacles are sampled according to realistic distributions. We generate a total of 7080 tasks by considering randomized initial and final state configurations for each task instance in the original ALFWorld dataset. Table 2 shows the task distribution according to each data split. As shown, the dataset contains seven task categories which vary in difficulty. The dataset is also partitioned into four subsets: *Train* which contains the most of each type of task; *Validation* which shares the same set of scenes and task instances with *Train*, but with randomized object locations and appearances; *Test Seen* is similar to *Validation*, although it should not be used during the training of agents; *Test Unseen* which consists of entirely new tasks and scenes, but with possibly seen object-receptacle pairs. Since we will be evaluating classical planners on SGPlan, the expected effects of distribution shifts between the *seen* and *unseen* tasks and scenes should not impact planning performance. Although, we still leverage all samples to more comprehensively understand the benchmark, and identify whether the same task types are objectively more challenging for classical methods in the different subsets.

Table 3: Task descriptions and complexity for the seven categories in SGPlan.

| **Task** | **Difficulty** | **Description** |
|---|---|---|
| Examine object in Light (LoiL) | Easy | The agent is tasked with locating and picking up an object of a specified class, before transporting it to a nearby light source. The task is complete when the light source is toggled and the object is held by the agent. |
| Pick and Place (PaP) | Easy | The agent must identify the location of an object with a specific class, pick it up, and transport it to a specified goal location. The task is complete when the agent has dropped the object on or in the target receptacle. |
| Stack and Place (SaP) | Medium | The agent must identify the location of two objects with specific classes: a non-receptacle object, and a receptacle object in which the first object can be placed (stacked) on. The task is complete when the objects are stacked and located in or on the target receptacle. |
| Pick two and Place (P2P) | Medium | The agent must identify the location of two objects which may be receptacles or non-receptacles, pick them up, and place each of them in their target receptacle locations. The task is complete when each object is located in its target location. |
| Pick Cool and Place (PCoP) | Hard | The agent is tasked with finding an object of a specified class, picking it up, cooling it, and placing it in a target receptacle. The task is complete when the object is cooled (object state) and located in the target receptacle. |
| Pick Heat and Place (PHeP) | Hard | The agent is tasked with finding an object of a specified class, picking it up, heating it, and placing it in a target receptacle. The task is complete when the object is heated (object state) and located in the target receptacle. |
| Pick Clean and Place (PClP) | Hard | The agent is tasked with finding an object of a specified class, picking it up, cleaning it, and placing it in a target receptacle. The task is complete when the object is clean (object state) and located in the target receptacle. |

In general, all tasks require interaction, at the symbolic level, with at least one object and receptacle in the scene. A high-level description for each task is given in Table 3, along with their relative difficulty. We develop an experimental testbed for rapidly benchmarking classical planners on SGPlan. Denoting the set of all tasks by $\Omega_a$, we say that this set is the union of solvable and unsolvable tasks: $\Omega_a = \Omega_s \cup \Omega_{\neg s}$. Unsolvable tasks can be regarded as redundant;

they are ill-posed problems generated by the ALFWorld task sampler, where the goal condition either equates to false or true at the initial state, or the goal condition contains an object not declared in the problem file. Executing a planner on these problems will result in a process error, and hence, they are not accounted for when computing primary and secondary performance metrics for planners. Although, we track $|\Omega_{\neg s}|$ across planners as a sanity check on the condition of SGPlan. As suggested by Rearrangement [18], we compute the task completion rate as the primary metric for evaluating the capability of planners:

$$TC_{\pi} = \frac{|\pi(\Omega_s)|}{|\pi(\Omega_s)| + |1 - \pi(\Omega_s)|}$$

Here, $\pi$ denotes a task planner, and $\pi(\Omega_s)$ is an output set, where $\pi(\Omega_s)_i = 1$ if a solution to $i$-th problem is found, or $\pi(\Omega_s)_i = 0$ if the planner failed to find a solution or timed out. To gauge the efficiency and optimality of planners and their computed trajectories, we also track the average planning time (wall-clock time) and the average plan length, respectively. Let $\Omega_{\pi,t} = \{t_i\}, \forall i \; iff \; \pi(\Omega_s)_i = 1$ be the set of plan times for all solved tasks by planner $\pi$. Similarly, let $\Omega_{\pi,\ell} = \{\ell_i\}, \forall i \; iff \; \pi(\Omega_s)_i = 1$ be the set of solution lengths for all solved tasks by planner $\pi$. Then, then we compute the plan time ($T_\pi$) and plan length ($L_\pi$) metrics as follows:

$$T_{\pi} = \frac{\sum_j \left(\Omega_{\pi,t}\right)_j}{|\Omega_{\pi,t}|} \quad L_{\pi} = \frac{\sum_j \left(\Omega_{\pi,\ell}\right)_j}{|\Omega_{\pi,\ell}|}$$

A potential limitation of SGPlan is that it currently does not incorporate a *do no harm* constraint on the domain and task instances. While there are a few potential avenues for integrating such a constraint in future iterations of the benchmark, the difficulty lies in developing a test that is compatible with all planners. For instance, PDDL [24] defines safety constraints as a viable approach to enforcing logical conditions over specific objects in a domain at the end of a task, such as *all receptacles of type fridge must be closed* or *all toggleable receptacles must be off.* Solutions that violate the hand-crafted safety constraints are invalidated, achieving the desired *do no harm* constraint. However, safety constraints must be self-imposed, meaning the planners themselves must implement checks on domain. Unfortunately, this feature is rarely supported by planners, and hence, safety constraints are not a scalable approach to incorporating *do no harm*.

We conclude that a *do no harm* constraint will likely require a plan validation tool for the environment, which we aim to introduce in later versions of SGPlan.

### 3.1.2 Benchmarking Classical Planners

We begin by formalizing the planning problem that planners will attempt solve in SGPlan, making note that tasks in the benchmark are described with their own PDDL problem file, but share the same domain file. The domain file specifies the set of objects types in the environment, $\mathcal{O}$, the set of predicates over these objects, $\mathcal{P}$, and the action space for the planner, $\mathcal{A}$. Each task instantiates a number of interactable objects $o \in \mathcal{O}$, an initial state $\mathcal{I}$, and a goal state $\mathcal{G}$. The initial state of the problem is defined in terms of an assignment of values to predicts $p_{\mathcal{I}} \in \mathcal{P}$ over objects $o$. In SGPlan, these initial logical predicates define the hierarchical support structure of the scene along object states. The planning domain consists only of deterministic state transitions, and hence, the actions from a policy or planner $\pi : \mathcal{S} \mapsto \mathcal{A}$ solely determine the assignment or deallocation of values to properties and predicates over objects, e.g., $\mathcal{T}(s_t, a_t) = s_{t+1}$ with probability one. As one might observe based on the task definitions in Table 3, the goal state $\mathcal{G}$ for each task type is conditioned on just a few object states, $p_{\mathcal{G}}$ over $o_{\mathcal{G}} \subseteq o$, rather than the full scene state. The goal of a policy is to find a sequence of actions, which will we denote by $a_\pi = \{a_1, a_2, \dots, a_m\}, a_i \in \mathcal{A}$, that drives the initial state $\mathcal{I}$ of the scene into the goal configuration $\mathcal{G}$, or to report that no such sequence of actions exists.

Table 4 shows the discrete action space of the policy in the SGPlan domain. Notice that the *go-to-location* action confines the agents position to receptacle locations only, which correspond to *places and structures* layer in the 3D scene graph proposed by *Rosinol et al.* [5]. However, since each scene corresponds to a single room, the agent in unable to move throughout broader levels of the scene graph, which may limit the horizon of realistic tasks in SGPlan. In addition, while action costs are defined, they remain unused by planners in search of a solution. Thus, the cost of an action sequence will only signify the types of actions a planner takes to achieve a particular task, which is likely proportional to the number actions taken (plan length). Given that SGPlan inherited the costs from ALFWorld, and the hand-crafted action operators could have been tailored towards low-cost solutions, we opt out of tracking the cost metric in our experiments.

Table 4: Action space for the planning problems in SGPlan.

| ACTION ($a \in \mathcal{A}$) | PARAMETERS ($o_a \in o_t$) | PRECONDITION ($p_t^{o_a} \in p_t$) | EFFECT ($p_{t+1}^{o_a} \in p_{t+1}$) | COST |
|---|---|---|---|---|
| **LOOK** | Agent-a, Location-l | AtLocation(a,l) | Examined(l) | 0 |
| **EXAMINE-RECEPTACLE** | Agent-a, Receptacle-r | ⋀(Exists(l), AtLocation(a,l), ReceptacleAtLocation(r,l)) | Checked(r) | 0 |
| **GO-TO-LOCATION** | Agent-a, Location-s, Location-e, Receptacle-r | ⋀(AtLocation(a,s), ReceptacleAtLocation(r,e)) | ⋀(~AtLocation(a,s), AtLocation(s,e)) | 1 |
| **OPEN-OBJECT** | Agent-a, Location-l, Receptacle-r | ⋀(Openable(r), ~Opened(r), AtLocation(a,l), ReceptacleAtLocation(r,l)) | ⋀(Opened(r), Checked(r)) | 1 |
| **CLOSE-OBJECT** | Agent-a, Location-l, Receptacle-r | ⋀(Openable(r), Opened(r), AtLocation(a,l), ReceptacleAtLocation(r,l)) | ~Opened(r) | 1 |
| **PICKUP-OBJECT** | Agent-a, Location-l, Object-o, Receptacle-r | ⋀(Pickupable(o), AtLocation(a,l), ReceptacleAtLocation(r,l), InReceptacle(o,r), ~HoldsAny(a), ⋁(~Openable(r), Opened(r))) | ⋀(~InReceptacle(o,r), Holds(a,o), HoldsAny(a), ~ObjectAtLocation(o,l) ) | 1 |
| **PUT-OBJECT** | Agent-a, Location-l, Object-o, Receptacle-r, OType-ot, RType-rt | ⋀(ReceptacleType(r,rt), ObjectType(o,ot), CanContain(rt,ot) AtLocation(a,l), ReceptacleAtLocation(r,l), Holds(a,o), ⋁(~Openable(r), Opened(r))) | ⋀(InReceptacle(o,r), ~Holds(a,o), ~HoldsAny(a), ObjectAtLocation(o,l)) | 1 |
| **CLEAN-OBJECT** | Agent-a, Location-l, Object-o, Receptacle-r | ⋀(Cleanable(o), Holds(a,o), AtLocation(a,l), ReceptacleAtLocation(r,l), ReceptacleType(r,SinkBasinType)) | IsClean(o) | 5 |
| **HEAT-OBJECT** | Agent-a, Location-l, Object-o, Receptacle-r | ⋀(Heatable(o), Holds(a,o), AtLocation(a,l), ReceptacleAtLocation(r,l), ReceptacleType(r,MicrowaveType)) | ⋀(~IsCool(o), IsHot(o)) | 5 |
| **COOL-OBJECT** | Agent-a, Location-l, Object-o, Receptacle-r | ⋀(Coolable(o), Holds(a,o), AtLocation(a,l), ReceptacleAtLocation(r,l), ReceptacleType(r,FridgeType)) | ⋀(IsCool(o), ~IsHot(o)) | 5 |
| **TOGGLE-OBJECT** | Agent-a, Location-l, Object-o, Receptacle-r | ⋀(Toggleable(o), AtLocation(a,l), ReceptacleAtLocation(r,l), InReceptacle(o,r)) | ⋀(Cond(IsOn(o) → ~IsOn(o)), Cond(~IsOn(o) → IsOn(o)), IsToggled(o)) | 5 |
| **SLICE-OBJECT** | Agent-a, Location-l, Object-co, Object-ko | ⋀(Sliceable(co), AtLocation(a,l), ObjectAtLocation(co,l), Holds(a,ko), ⋁(ObjectType(ko,KnifeType), ObjectType(ko,ButterKnifeType))) | IsSliced(co) | 5 |

For the initial benchmark, we consider two state-of-the-art classical planners, Fast Forward X [20] and Fast Downward [21], as they have demonstrated efficiency for solving long-horizon planning problems at the IPC [26] and in modern planning environments [11, 30, 106]. They have also been used as competitive baselines in robot task planning works [7, 12]. Furthermore, we include variants of the FD-plan system, namely, FD-plan with Dijkstra's (FD-Dijkstra's) to assess the speed trade-off when attempting to solve for the shortest path trajectory, and FD-plan with Context Enhanced Additive (FD-CEA) [108] to evaluate the effect of improved heuristics on planning efficiency. While there are other performant planners of interest such as UPMurphi [54] and DiNo [9], they are not compatible with SGPlan, due to a lack of support for existential clauses which are extensively used in problem files. In fact, in order to run FF-plan, which was originally designed for standard PDDL domains, we further modified the domain file to remove the dependency on numeric fluents (action costs) and stiffened the requirements on object type specificity in action preconditions defined in the original domain file. Even with these modifications, we utilized an extended version of FF-plan system called FF-X, which supports a subset of PDDLv.2.2 features required in SGPlan. The compatibility challenges faced across planners, with the exception of FD-plan which the original domain was designed for, are indicative of an over-engineered domain structure. Our hypothesis is that because data generation was the intended purpose of task planners in the original ALFWorld domain, little effort was made to ensure that the PDDL domain was compatible with alternative planners. Nonetheless, the above modifications are sufficient for our purposes; namely, benchmarking two of the most commonly used classical planners [20, 21] on wide-range of robot task planning problems in hierarchical scenes.

## 3.2 PDDLGym Planning with Graph Neural Networks

The original design of PLOI [7] aimed to capture relational information inherent to a problem instance in order to prune redundant regions of the state space, and thereby create an abstracted problem where planning is more efficient. As an instantiation of this concept, referred to as projective abstraction, they apply a standard GNN model [109] to process an input graph, where nodes and edges correspond to objects and the logic-based relationships between them, respectively. They condition the GNNs' inference on the goal state by also embedding final state edge attributes between objects in the input graph. They proceed to classify the importance of objects with respect to the task, threshold the importance scores to acquire a reduced subset of objects, and attempt to plan. If a plan is not found, the threshold is decreased and a larger subset of objects is retained for planning – the process repeats until a viable solution is found. The GNN model is weakly supervised by sufficient object sets obtained by iteratively and greedily reducing the number of objects in the planning problem until FF-plan fails to find a solution. This strategy addresses the trend shown in Figure 12, emphasizing the exp-time effect of extraneous objects on planners. Therefore, the expected effect of projecting-out a conservative number of these irrelevant objects could yield significant savings in plan time.

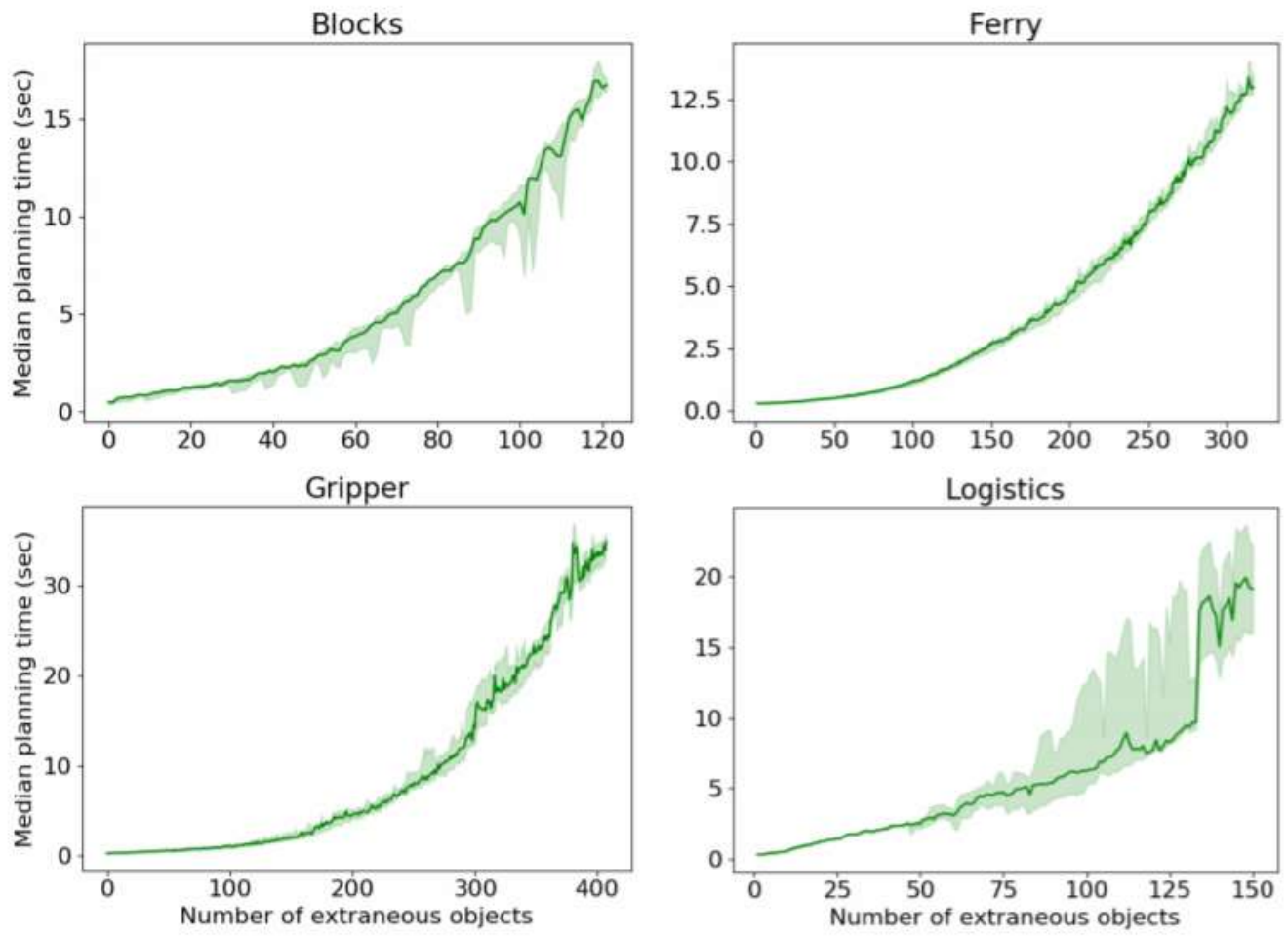

Figure 12: The effect of extraneous objects on planning time [7].

This framework is applied to hallmark classical planning problems available in PDDLGym [30], and has shown impressive results compared to RL-based policies, modern lifted planners, and learning-based action grounding approaches. Of the experiments conducted, no explicit attempt was made on hierarchical environments, such as those contained in SGPlan. However, several of the classical planning domains share similar properties: predicates that reflect the physical support structures of the domain. These support structures are often hierarchical, establishing a correspondence between them and 3D scene graphs. Examples of these include Blocks, Towers of Hanoi, and the stochastic Exploding Blocks domain. As a result of having a much sparser state space, the Blocks-type problems more largely correspond to planning in scene graphs. In contrast, one can observe that Towers of Hanoi tasks require interaction with almost all objects, rendering the prediction of object importance scores rather needless. Therefore, we isolate our learning-to-plan experiments to the Blocks domain with the purpose of identifying the weak-points of PLOI and determining the appropriate extensions required for scene graph planning.

### 3.2.1 Sensitivity Analysis of PLOI

We design a sensitivity analysis on PLOI with the objective of acquiring a deeper understanding of the GNN model's behaviour with respect to the imperfect supervision quality that it is trained on. In this regard, one of our main concerns is with respect to the interpretability of the model's predictions, and the challenges this presents when attempting to tune the object importance threshold accordingly. Since sufficient object sets are generated through greedy uniform sampling and are validated by a planner of choice, the potential accuracy of the model is upper-bounded in two dimensions. First, in large problem instances, there is an extremely low probability that a minimum object set will be found through random sampling. Second, the validating planner may be unable to solve challenging problems, even with redundant objects removed, which limits the hypothesis space of tasks that will be observed by the GNN model during training. Thus, the trained GNN model may be susceptible to out-of-distribution tasks. Moreover, a hand-selected threshold value does not account for uncertainty in the model's predictions, and may lead to several failed planning attempts. As discussed in Section 2.3, the cost of computing heuristics is extremely costly. Thus, repetitive replanning should be avoided if possible, as it can accrue more cost than the inference time of the GNN model.

Since the training supervision from the GNN contains an element of stochasticity due to the uniform random dropping of objects to form sufficient object sets, we compare the trained GNN to a stochastic baseline which uniformly samples objects to be part of the planning problem. This baseline can be interpreted as a random guidance module or a uniform random importance scorer. In comparing the GNN model to this baseline, we highlight the extent to which learned inductive biases can cut through stochasticity in the training samples. Furthermore, we extend this experiment to investigate the GNNs' ability to rank the importance of nodes by sampling the top-K object scores and attempting to plan with them. We evaluate on ten challenging test problems which contain on the order of a hundred objects each, and with goal specifications over approximately twenty of those objects.

In order to analyze the model's sensitivity to supervision quality, we consider forming sufficient object sets of various sizes by terminating the greedy search early. For each problem instance in the Blocks domain, we collect labels from sufficient object sets with the top-50 objects and top-100 objects. The final training set contains a total of forty problems, and we evaluate the trained models on five test problems which contain on the order of a hundred objects each.

### 3.2.2 Graph Attention Networks and Relation Importance

The graph neural network introduced by *Silver and Chitnis et al.* [7] reflects a very standard message passing model. To illustrate the computation flow of their model, we adopt the graph networks formalism provided by *Battaglia et al.* [109]. Let us consider a graph that would like to process, defined by vertices $V = \{(\boldsymbol{v}_i)\}_{i=1:N}$ and edges $E = \{(\boldsymbol{e}_k, r_k, s_k)\}_{k=1:M}$, where $\boldsymbol{v}_i$ and $\boldsymbol{e}_k$ are node and edge attributes, and $(r_k, s_k)$ are the receiver and sender node indices for edge $e_k$. In the case of PLOI, node attributes correspond the semantic class of the object and any unary predicates of the initial state or goal state. Likewise, edge attributes correspond to binary predicate relationships between objects in the initial state or goal state (see Figure 9). PLOI's GNN architecture can be defined by two update functions, $\phi$, and one permutation invariant set aggregator function, $\rho$:

$$\boldsymbol{e}'_k = \phi^e(\boldsymbol{e}_k, \boldsymbol{v}_{rk}, \boldsymbol{v}_{sk}) = NN_e([\boldsymbol{e}_k, \boldsymbol{v}_{rk}, \boldsymbol{v}_{sk}])$$

$$\bar{\boldsymbol{e}}'_i = \rho^{e \to v}(E'_i) = \frac{1}{|E'_i|} \sum_j \boldsymbol{e}'_j$$

$$\boldsymbol{v}'_i = \phi^v(\boldsymbol{v}_i, \bar{\boldsymbol{e}}'_i) = NN_v([\boldsymbol{v}_i, \bar{\boldsymbol{e}}'_i])$$

Here, $E'_i = \{(\boldsymbol{e}_k, r_k, s_k)\}_{rk=i,k=1:M}$, is the subset of incoming edges to receiver node $rk$. When analyzing the class of planning problems that PLOI targets, we observe that the number of edge types are rather limited, and that nodes in the graph have a fairly consistent quantity of in-degrees and out-degrees. As a result, their proposed architecture, which allows edges embeddings to equally contribute in the permutation invariant aggregation function, $\rho$, suffices in these classical planning domains.

In transitioning to 3D scene graphs, we would like to design a more expressive GNN model that is capable of reasoning over the feature rich node and edge attributes of the scene. In particular, edges in the scene graph will not only correspond to interlayer support relations, but will also describe intralayer relationships such as co-visibility and object affordances. Conditioned on a task, these edge embeddings should not be treated as equal in the edge aggregation step ($\rho$). For instance, when attempting to search for an object in a 3D scene graph, co-visibility is a far more informative edge attribute than object affordance. On the contrary, for the task of preparing a meal which involves object-to-object interaction, affordances become increasingly important. Considering the wide range of tasks available in SGPlan, we require a GNN model with sufficient ability to weight edge embeddings according to their relevance for a task. Therefore, we modify PLOI's architecture with self-attention, and propose to use a graph attention network [75] which closely follows the non-local neural network model proposed by *Wang et al.* [110]:

$$\boldsymbol{e}'_k = \phi^e(\boldsymbol{e}_k, \boldsymbol{v}_{rk}, \boldsymbol{v}_{sk}) = \left(\phi^a(\boldsymbol{v}_{rk}, \boldsymbol{v}_{sk}), \phi^u(\boldsymbol{v}_{sk}, \boldsymbol{e}_k)\right) = (\alpha'_k, \boldsymbol{b}'_k)$$

$$\bar{\boldsymbol{e}}'_i = \rho^{e\to v}(E'_i) = \frac{1}{\sum_k \alpha'_k} \sum_j \alpha'_j \boldsymbol{b}'_j$$

$$\boldsymbol{v}'_i = \phi^v(\boldsymbol{v}_i, \bar{\boldsymbol{e}}'_i) = NN_v([\boldsymbol{v}_i, \bar{\boldsymbol{e}}'_i])$$

The contribution of our model is two-fold. First, edge attention increases the expressivity of the GNN and should allow for more descriptive embeddings to be extracted from the input graph, particularly as we transition the architecture to scene graph style domains. Our second insight is that by modelling the importance of edges with respect to a task, we can begin to reason about their relevance. Conceptually, this extends PLOI by allowing the GNN to project-out irrelevant relationships between objects, as well as the objects themselves, as depicted in Figure 13. From an implementation perspective, our GNN is now able to predict importance scores for edges (predicates) of the planning problem, and with an independent relation threshold, either retain or project them out. The advantage of our method is that predictions are now made at a more granular level, as opposed to predicting which objects to retain, and in turn retaining all of their associated edges. As an example, a heavy piece of furniture needs to be moved in a house rearrangement task. The relation-level predictions of our GAT could introduce the distinction between a plan which considers lifting the heavy piece of furniture versus sliding it across the floor. This level of control was not previously achievable by the model used in PLOI.

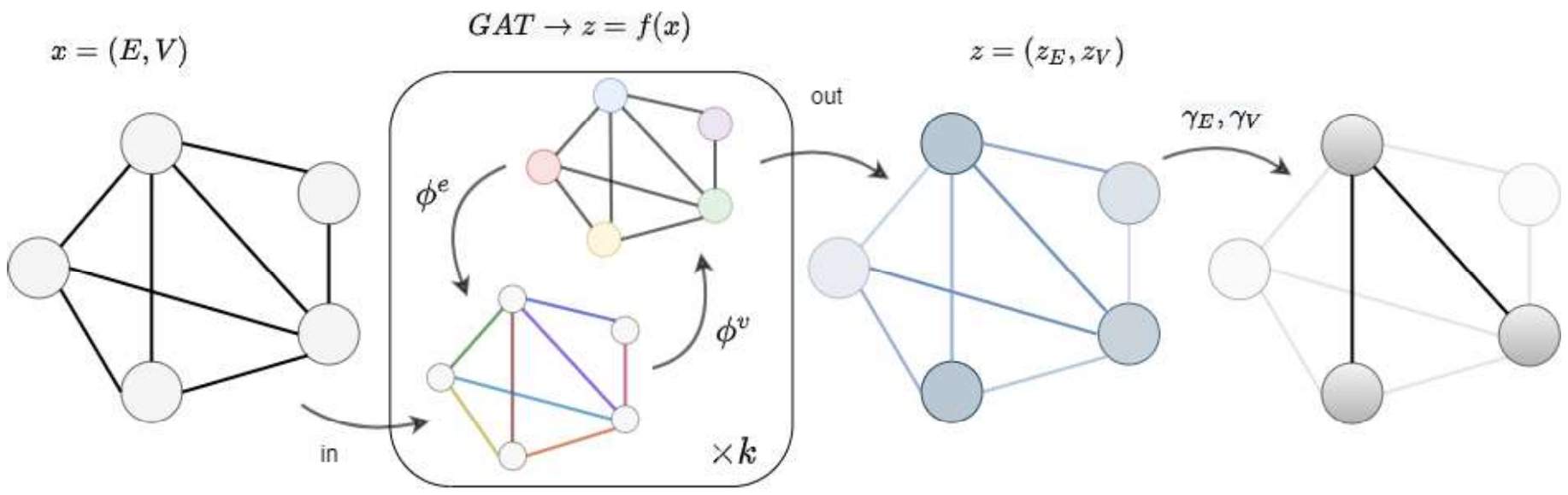

Figure 13: Graph attention network [75] for object importance and relation prediction.

Upon these changes, we do not anticipate that our graph attention network will have a significant performance boost on the quality and efficiency of planners in the Blocks domain, due to the uniformity in edge attributes. However, we suggest that this modification, amongst more to come, is a necessary step towards learning meaningful scene graph representations for downstream planners. Future considerations include explicit modelling of edge types with relational graph neural networks [73], and combining the benefits of self-attention with multi-relational networks [76].

### 3.2.3 Sufficient Object Sets via Regression Planners

In this section, we address the concern of supervision quality used to train our graph neural network models. As with most machine learning and artificial intelligence applications, the quality of supervision matters; recall that the PLOI model uses sufficient object sets to train the GNN model as an object scorer. For a single planning problem, we recall the process used to generate binary training labels for all objects in the problem:

1. Let $O = O_+ = \{o_1, o_2, \ldots, o_n\}$ be the current object set, initialized to the full set of objects.
2. Randomly sample an object to remove: $O'_+ = \{O \setminus o_k\}$, $k \sim u(1, n)$.
3. Execute FF-plan on the abstracted planning problem.
4. If a solution is found, let $O_+ = O'_+$, and repeat from step 2.
5. If no solution is found, return $O_+$ and $O_- = \{O \setminus O_+\}$.

It is apparent that this stochastic sampling process while likely generate imperfect training data, even if the procedure is repeated several times per problem in order to coincidentally construct smaller object sets. While one can argue that data generated in this way is diverse, thereby preventing model overfitting, we would emphasize that there are better alternatives than to achieve diversity at the cost of introducing noise into our dataset.

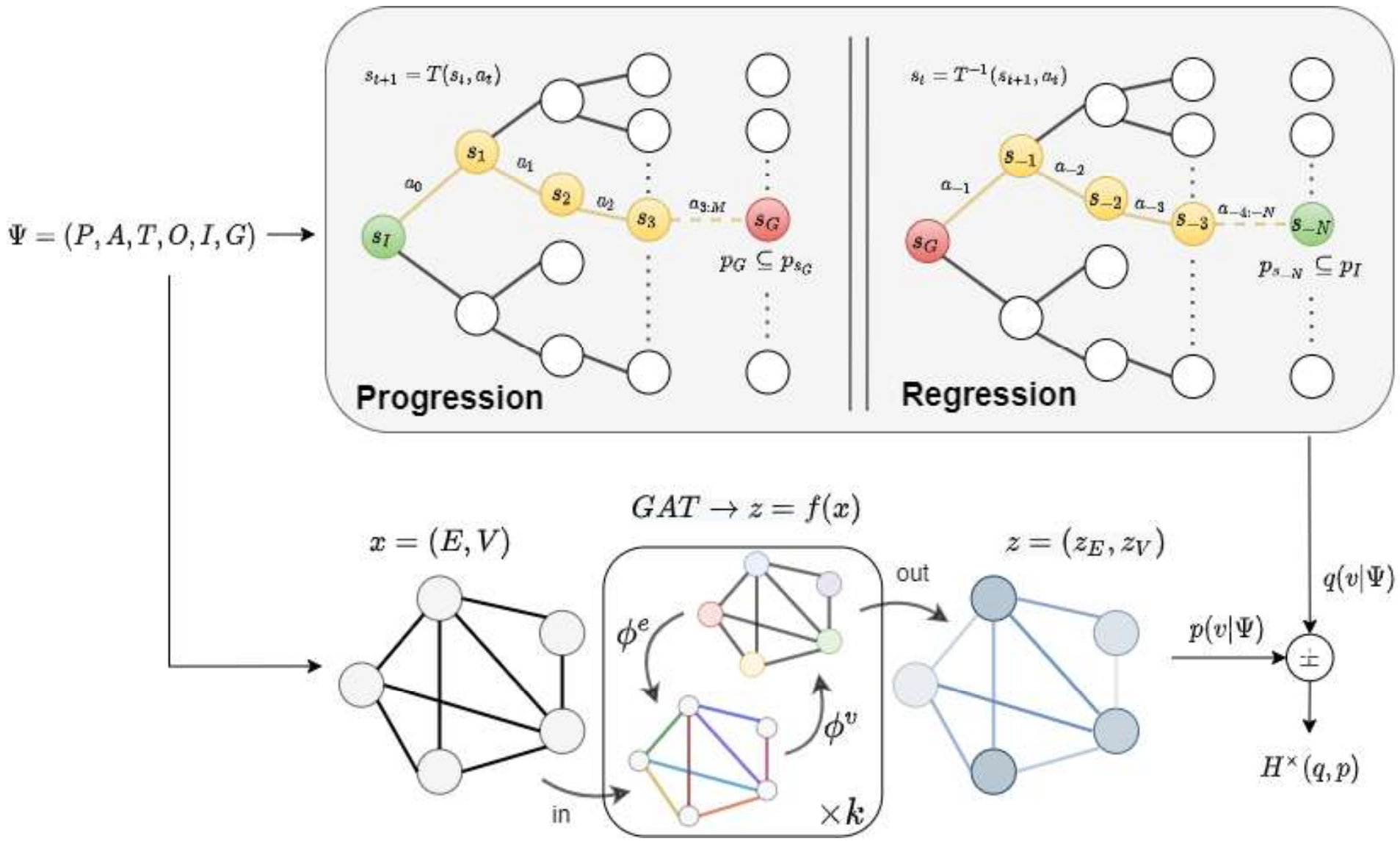

Figure 14: Leveraging regression planners to supervise training of the GAT model.

We propose an alternative strategy to generating object-level class labels. In particular, we make use a regression planner [13] to generate a diverse dataset of trajectories. For a given task, we iterate through the sequence of states generated by the regression planner, and append each object involved in transitions along the reverse trajectory to the sufficient object set, $O_+$. All other objects in the domain are assigned a negative label. Through this procedure, important objects correspond to those that lie directly on a trajectory regressing from the goal state, and hence, these objects directly contribute to achieving the goal. Our notion of object-level importance is far stronger than the previous, which considered any object as part of a plannable set to be important for a planning problem. However, one potential concern in our approach to generating supervision, despite providing the GNN with more informative labels, is that it will yield predictions that are less conservative compared to PLOI. As previously mentioned, misclassifications in the PLOI framework could be very costly, as it will require several iterations of expanding the object set and replanning.

To assess to the precision of our GAT model trained with regression-based sufficient object sets, we limit the number of planning iterations to one. Hence, we obtain a one-shot model and empirically tune the object importance threshold. We conclude by comparing the results of this model to a one-shot implementation of the PLOI model.

### 3.2.4 Spatial Edge Attributes

As PLOI has shown, the initial and goal states of classical planning domains can be casted into a relational representation, where relations reflect binary predicates over objects in the domain. In three message passing iterations, the GNN model must leverage this relational information to acquire a sufficient *understanding* of the problem and accurately infer which objects will require interaction in the process of solving a planning problem. As we know, the k-hop neighborhood size embedded in the final node features grows exponentially with the number of layers in a graph neural network, similar to how the receptive field size of a neuron grows with additional convolutional layers in deep CNNs. After three message passing iterations, each embedding will reflect its 4-hop neighborhood. In smaller domains, 4-hop neighborhood features might capture enough of the problem structure to decide whether or not a node is important. Although, in larger domains where the solution space of a planner has more coverage of the input graph, we may need the final node embeddings to also span across larger k-hop neighborhoods. To this end, the number of layers in a graph neural network is a domain-specific hyperparameter that PLOI has selected to suit the size of the classical planning domains and their corresponding graphical structures. This imposes a limitation on the size of planning problems that we could expect the GNN to perform well on, particularly if the model has only observed smaller training instances.

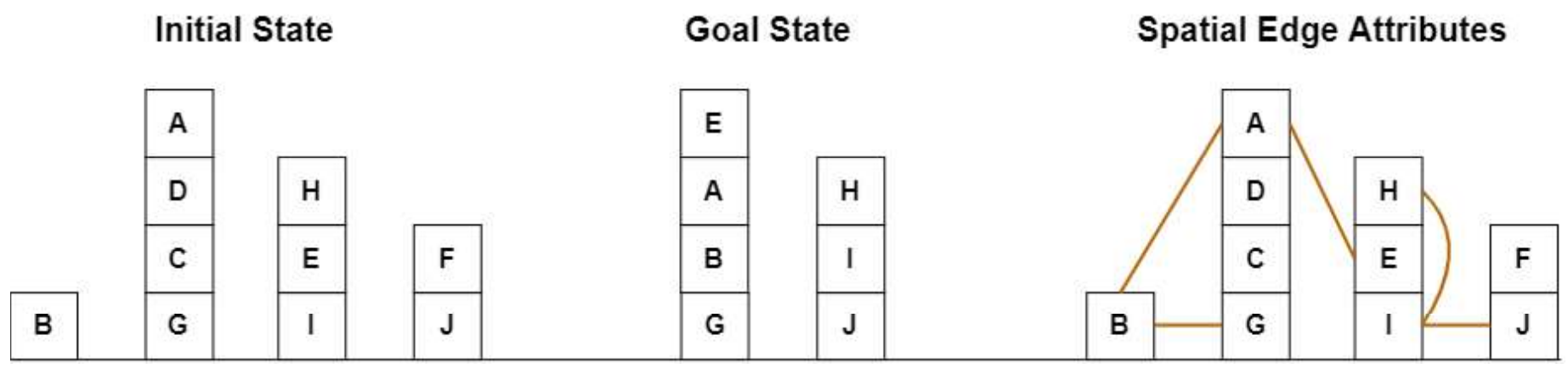

Figure 15: Spatial edge attributes in PDDLGym Blocks domain.

An as alternative to increasing the depth of the GNN model, which would be beneficial, we anticipate that appending simple but informative node and edge attributes could improve prediction quality. Moreover, 3D scene graphs are feature rich; the ability to harness these attributes was the primary motivation for adopting the GAT network as discussed in the previous section. When composing relational representations for classical planning domains, PLOI only considers the support relations of the problem. These support relations represent the minimal relational structure of the domain, much like the interconnections between layers in a hierarchical scene graph. Thus, for all edges in the Blocks planning problem, we compute a Manhattan distance between the connected blocks and append the metric as a spatial edge attribute, as shown in Figure 15. In doing so, the GNN no longer needs to extract an implicit spatial representation of the domain from the raw support relations. Conceptually, spatial attributes, in combination with support relations, can act as an approximate heuristic on the number of steps or actions required to bring two or three blocks into their relative goal state configurations from the initial state. In demonstrating that the GNN could exploit spatial relationships for more accurate predictions, we can gain insight on the expected performance of the model in 3D scene graphs, which in addition to spatial information contain visual, physical, semantic, and affordance relations between objects.

# Section 4: Results and Discussion

## 4.1 Summary of Results: Classical Planning in SGPlan

After generating planning problems in PDDL format as per the task distribution given in Table 2, we run experiments over the entire benchmark with state-of-the-art classical planners: Fast Forward X [20], Fast Downward Classic [21], Fast Downward with Context Enhanced Additive [108], and Fast Downward with Dijkstra's Algorithm. We note that the PDDLv2.2 domain used in these experiments was originally written for the features supported by FD-plan, but received modifications in order to make the planning problems compatible with FF-X.

For each planner, we track task completion as the primary evaluation metric, with plan time and plan length as secondary metrics. In order to fully understand the strengths and limitations of the benchmark, we track these metrics at various levels. By averaging the results over the data splits (i.e., *Train, Validation, Test Seen, Test Unseen*), we can empirically determine which of the splits contain more challenging problems on average. Alternatively, by average the metrics across task categories, we can identify which task types impose more state changes, and draw correlations between plan length and plan time. At the most granular level, we aim to assess the difficulty of the same task categories in different partitions of the dataset. An in depth understanding of the dataset is important as we transition to benchmarking neural symbolic planners and other learning-based methods. Unlike classical planners, these online planners can be highly sensitive to distribution shifts across splits and task categories. Thus, in addition to discerning whether SGPlan is a viable benchmark for scene graph planning, we will also be equipped to draw appropriate conclusions from learning-to-plan techniques going forward.

Table 5: Task completion rate ($TC_{\pi}$) of classical planners across SGPlan task categories.

| | *LoiL* | *PaP* | *SaP* | *P2P* | *PCoP* | *PHeP* | *PClP* | $\mu$ | $\sigma$ |
|---|---|---|---|---|---|---|---|---|---|
| *FF-X* | 0.02 | 0.16 | 0.00 | 0.16 | 0.00 | 0.00 | 0.05 | 0.06 | 0.07 |
| *FD-Classic* | 1.00 | 1.00 | 1.00 | 1.00 | 1.00 | 1.00 | 1.00 | 1.00 | 0.00 |
| *FD-Dijkstra's* | 1.00 | 0.99 | 1.00 | 0.86 | 0.77 | 0.67 | 0.89 | 0.88 | 0.12 |
| *FD-CEA* | 1.00 | 1.00 | 1.00 | 1.00 | 1.00 | 1.00 | 1.00 | 1.00 | 0.00 |

A summary of the success rates across task categories is given in Table 5. It is clear that FD-plan and its variants are able to consistently compute solutions within the 10 second timeout, while FF-X struggles to. Unfortunately, FF-X very rarely solves even the simplest of task categories, such as pick-and-place and stack-and-place. As we would expect, the overhead of Dijkstra's shortest path algorithm when incorporated in Fast Downward leads to more timeouts, particularly in the long-horizon regime (i.e., pick-heat/cool/clean-and-place). There are several potential causes for the observed performance gap between FF-X and FD-plan. First, such robot planning problems might contain strong causal dependencies: e.g., the problem can be decomposed into task hierarchies where the state changes of some objects are causally dependent on others. Under this assumption, FD-plan is expected to outperform most classical planners, as it transforms the propositional PDDL or STRIPS like representation into causal dependency graphs, and informs the search expansion in this domain with its causal graph heuristic [21]. Indeed, the state changes of all objects in the scene graph depend on the location of the robot agent, similar to the Logistics classical planning domain.

In contrast, the enforced hill-climbing strategy of FF-X simply commits to one path of search tree in its standard propositional representation, as an exhaustive search of the state space would be computationally intractable. Despite the improvements made over HSP [19], the GraphPlan [52] based heuristic utilized in FF-X is still inadmissible, and hence, the planner can only guarantee satisficing solutions. Should the enforced-hill climbing strategy greedily commit to a promising path in the search space which does not contain a solution, the planner is likely to timeout. Another possibility concerns the conditioning of the problems after modifying the domain to make it PDDLv2.1 compatible for FF-X. While highly unlikely, it is possible that these modifications do not permit certain facts to be generated in the search process, yielding dead-ends in the search space. However, considering that FF-X managed to solve tasks involving object state changes and toggleable receptacles, such as pick-clean-and-place and look-at-object-in-light, we believe that the high failure rate of FF-X is indicative of the algorithms incapacity rather than a product of ill-conditioned tasks.

We provide qualitative examples of trajectories generated by FF-X and FD-plan in Figure 16. The top row illustrates one of the few successful pick-heat-and-place task instances for the FF-X planner. The bottom row shows a more challenging version of the pick-heat-and-place task which requires the object's state to be changed, in this case sliced with a knife, before being heated and placed in a target receptacle. We find that both the classic and CEA variant of the FD-plan system are able to consistently solve these long-horizon tasks, while the Dijkstra's variant begins to timeout more frequently with increasing task complexity. For additional visualizations of trajectories generated by these planners, and for solutions that would violate a *do no harm* constraint, we refer the reader to Figures 23-24 in Appendix. A.

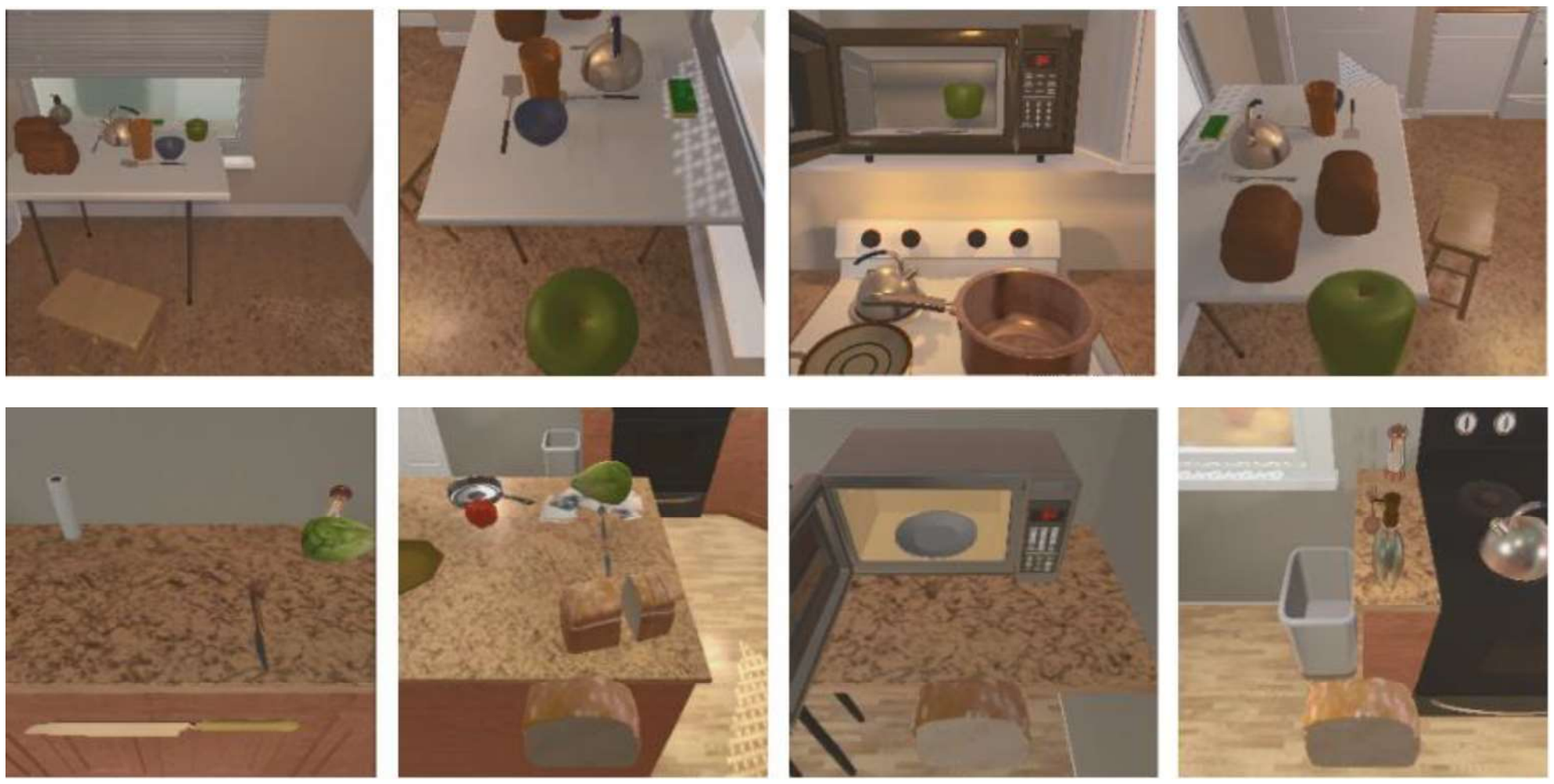

Figure 16: Visualization of successful pick-heat-and-place tasks from FF-X (top) and FD-plan (bottom).

We conduct a series of experiments to investigate the plan time and plan lengths produced by the four planners across the task categories and splits of SGPlan; the results are given in Figure 17. Recall that these metrics are computed only for successful plans, and do not account for timeouts or failed attempts. Thus, as per the task completion rates shown in Table 5, these metrics are averaged over a much smaller solution set for FF-X compared to the FD-plan based systems. We begin by analyzing results over task categories (left column). Across all task categories, FF-X requires significantly more time to find solutions compared to other benchmarked planners. Specifically, for the look-at-object-in-light task, FF-X solutions are over twenty times more costly to compute than the slowest Downward planner. Moreover, while the plan time of FD-planners resemble an upward trend across the ordered tasks, FF-X plan times reflect the inverse trend. This observation is quite intriguing, as it indicates that the tasks most easily solved by FD-plan are the ones that pose the most difficulty for FF-X. However, this claim does not account for the stack-and-place (SaP), pick-heat-and-place (PHeP), and pick-cool-and-place (PCoP) tasks, which FF-X is unable to solve in general.

Further analysis in the corresponding solution lengths demonstrate consistency between the plan times and plan length for the FD-plan system. This consistency was not observed for the FF-X planner; the task category that required the fewest state changes (look-at-object-in-light) was approximately two times more costly to solve in comparison to the task with the most state changes (pick-two-and-place). Although, in the set of mutually solved tasks by all of the planners, we observe consistency between the lengths of computed trajectories.

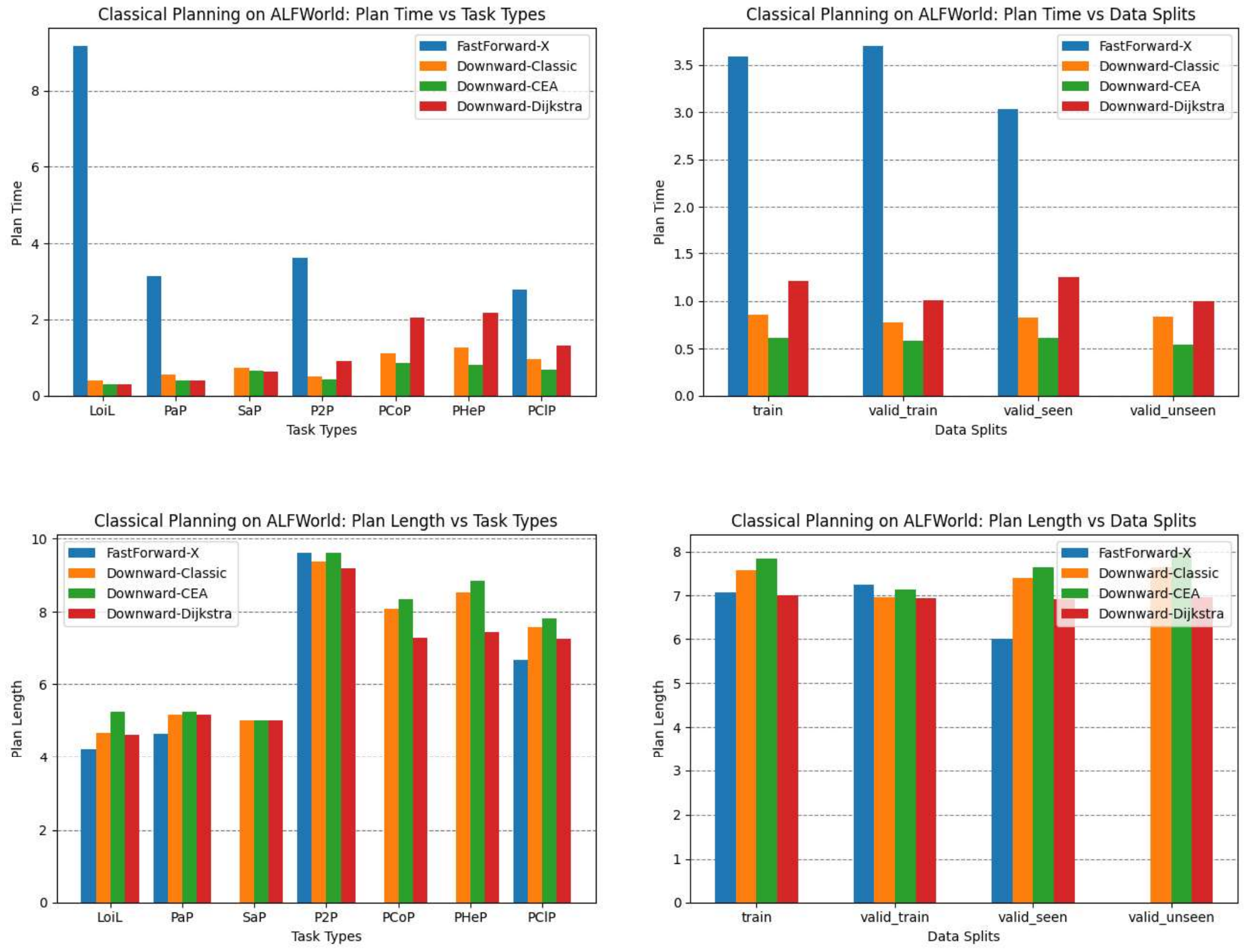

Figure 17: Plan time (top) and length (bottom) of classical planners across tasks (left) and splits (right).

The reason for FF-X's inability to solve the SaP, PHeP, PCoP tasks is unclear, especially since it solves the PaP and P2P tasks at equivalent frequencies, which together represent the extremes in the average solution length across all task categories. We hypothesize that FF-X struggles in tasks that demand interaction between multiple objects in the scene or repeated interaction with the same object. This constitutes action sequences such as *place fork in bowl*, *cool object in fridge*, or *heat object in microwave*. Unfortunately, these action sequences can be interpreted as bottlenecks in the search space which FF-X fails to locate, and hence, renders these tasks unsolvable in the allotted ten second runtime period. Once again, leveraging causal dependency, e.g., "the object can only be cooled when placed in the fridge", is perhaps the strategy that yields success for FD-plan in these contexts. Decomposing the binary task completion metric into the ratio of completed subgoals [18] achieved by the planner at the ten second timeout could potentially provide deeper insights on the limitations of the Fast Forward planning system, and upcoming learning-based planners which may also struggle in these long-horizon task settings.

We report identical metrics across the Train, Validation (*valid_train*), and Test sets (*test_seen*, *test_unseen*) of SGPlan in the right column of Figure 17. FD-plan shows consistent performance in plan time and solution length across all splits, where the Unseen Test set just slightly upper bounds the computed solution lengths with an average of eight steps. This indicates that the data splits are largely dominated by the task categories that demand more actions in order to solve. While this claim might seem contradictory to the uniform task distribution outlined in Section 3.1.1, we recall that a portion of all the tasks will be reported as unsolvable by the planners. In particular, FD-plan reports that approximately 30% of tasks in the Test Unseen set are unsolvable, where most of them correspond to the simpler look-at-object-in-light, pick-and-place, and stack-and-place tasks, as depicted in Figure 18. Uncoincidentally, FF-X is unable to solve a single task in the Test Unseen set, and takes about three times longer to find solutions in the remaining data splits. As observed across task categories, trajectories generated by all planners are of similar length, with FF-X and FD-Dijkstra's generating the shortest plans as expected. Interestingly, unsolvable tasks reported by FD-plan outnumbers those reported by FF-X by approximately ten percent. However, this might only account for a fraction of the performance variation between FF-X and its counterparts.

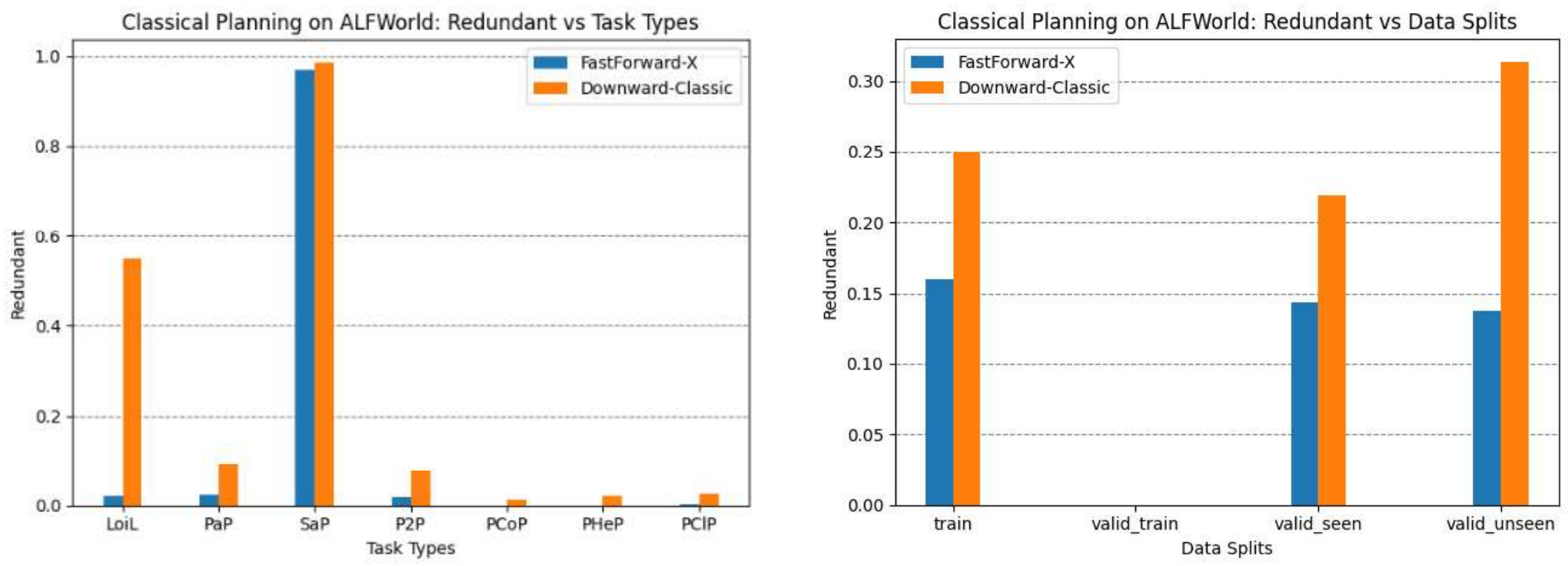

Figure 18: Number of reported unsolvable problems by planners across tasks (left) and splits (right).

Table 6: Task completion, plan time, and plan length of tasks across all SGPlan splits with FD-Dijkstra's.

| | *Train* | | | *Validation* | | | *Test Seen* | | | *Test Unseen* | | |
|---|---|---|---|---|---|---|---|---|---|---|---|---|
| | $TC_{train}$ | $T_{train}$ | $L_{train}$ | $TC_{val}$ | $T_{val}$ | $L_{val}$ | $TC_{seen}$ | $T_{seen}$ | $L_{seen}$ | $TC_{unseen}$ | $T_{unseen}$ | $L_{unseen}$ |
| *LoiL* | 1.00 | 0.29 | 4.63 | 1.00 | 0.30 | 4.58 | 1.00 | 0.30 | 4.54 | **<u>1.00</u>** | **0.32** | <u>4.78</u> |
| *PaP* | 1.00 | 0.30 | 5.16 | 1.00 | 0.30 | 5.15 | 1.00 | 0.32 | 5.21 | **<u>1.00</u>** | **0.44** | <u>5.58</u> |
| *SaP* | **<u>1.00</u>** | **0.65** | <u>5.00</u> | - | - | - | - | - | - | - | - | - |
| *P2P* | 0.86 | 0.94 | 9.18 | 0.98 | 0.63 | 9.18 | 0.88 | 0.85 | 9.11 | **<u>0.67</u>** | **1.43** | <u>9.57</u> |
| *PCoP* | **<u>0.76</u>** | 2.10 | 7.28 | 0.93 | 2.06 | 7.08 | 0.81 | **2.25** | <u>7.40</u> | 0.83 | 1.33 | 7.27 |
| *PHeP* | **<u>0.65</u>** | 2.21 | 7.40 | 1.00 | 1.90 | 7.52 | 0.76 | **2.30** | <u>7.72</u> | 0.81 | 1.53 | 7.44 |
| *PClP* | **<u>0.88</u>** | 1.35 | 7.24 | 0.95 | 1.27 | 7.23 | 0.89 | **1.49** | 7.27 | 0.91 | 0.72 | <u>7.32</u> |

We conduct an additional study which ablates the task completion, plan time, and plan length of FD-Dijkstra's across task categories in each data split, given in Table 6. We select the FD-Dijkstra's algorithm as it makes for an interesting combination of the shorter trajectory lengths produced by FF-X with a slight increase in cost from the classic FD-plan system. Notice that the majority of the stack-and-place tasks are invalidated by the planner, in correspondence with Figure 18. For all task categories, we observe that the largest plan times and plan lengths are associated with tasks in the Test Seen and Test Unseen data splits. However, the lowest task completion scores are attributed to problems in the Train set, with the exception of pick-two-and-place which falls in the Test Unseen set. Thus, in terms relative task difficulty between splits, we conclude that SGPlan is well-balanced. Although, it would be preferable for the test set to present a stronger challenge in order to gauge the generalizability of learning-based planners.

We summarize the findings of our benchmarking efforts on SGPlan as the following. While both the Fast Forward [20] and Fast Downward [21] planners are considered state-of-the-art in the classical planning literature, there are substantial computational advantageous in exploiting causal dependency graphs, as illustrated by the performance of FD-plan. The variants of FD-plan are performant alternatives which trade off solution length for reduced planning time (FD-CEA [108]) or pursue the converse trade-off (FD-Dijkstra's). Despite the quantity and distribution of tasks in SGPlan, the problems do not pose a significant challenge to the state-of-the-art FD-plan system, which achieves a task completion rate of one-hundred percent. In fact, as we will demonstrate in the upcoming sections, task instances in classical planning domains induce solutions that are an order of magnitude larger than those observed in the *long-horizon* tasks of SGPlan. Moreover, a significant portion of the problems generated through the ALFWorld [106] task samplers were reported as ill-posed; this concern of data quality prohibits any current initiatives for community release. Therefore, we see an opportunity to improve the benchmark along three directions.

- Task Complexity: Planning problems with longer horizons and a more realistic state space. In addition, we would like to investigate the feasibility of planning under partial observability, i.e., the planner only gains access to objects and associated logical facts once a region of the state space has been explored. In SGPlan, the agent is limited to moving between 3-5 receptacle locations, and predicates that govern motion and object interaction are carefully design so as to limit the branch factor of the search. While this facilitates efficient retrieval of high-level task demonstrations to train learning-based methods, it is not suitable for a benchmark intended to challenge state-of-the-art planners.
- Environment Scale: An AI2-THOR [31] based environment is limiting in the size and diversity of scenes. The scene graphs in SGPlan are constrained to a three-layer hierarchy, and encompass only a fraction of objects and receptacles accessible in household environments. Confining the Rearrangement [18] task to a single room will not foster research engagement at the intersection of scene graphs and planning. Thus, we believe that room for contribution lies in appropriately modifying the Gibson [101] environment to support high-level interaction at the scale of buildings.
- Metrics: More sophisticated evaluation metrics would enable richer comparisons to be drawn between classical and learning-to-plan methods. As mentioned, the two of interest includes a partial task completion metric and a *do no harm* metric. The partial task completion metric more fluidly evaluates the performance of the planner by accounting for the number of subgoals it completes. The *do no harm* metric would prevent unsafe or undesirable action sequences by invalidating the trajectory as a whole.

Thus, our findings from these preliminary benchmarking efforts will serve as a guide in the development (in progress) of a large-scale scene graph planning environment based on Gibson. This benchmark will compensate heavily on these three directives and will be uniquely tailored for research at the intersection of Rearrangement, Task Planning, and 3D Scene Graphs. In turn, we believe that its release will provide the robotics community with a fresh perspective on the importance of leveraging relational representations for long-horizon robot task planning.

## 4.2 Summary of Results: Planning with GNNs

We begin with an overview of the PDDLGym [30] Blocks classical planning domain, which is used extensively to analyze and develop improvements on the PLOI planning system [7]. In the Blocks problem, the initial state is defined by a set of blocks in some arbitrary configuration, and a solution is a sequence of actions that rearranges a subset of blocks into a specified goal state configuration. Interaction with blocks is limited to pick, place, stack, and unstack actions. This domain contains a hierarchical element in that object relations define a layered support structure, where objects can only be interacted with if they are leaves of the hierarchy (e.g., contain no objects placed on top of them). Thus, planning in Blocks shares an abstract relationship to planning in 3D scene graphs. In particular, this classical planning domain may very well pose a more difficult challenge compared to the tasks in SGPlan, as the hierarchical support structure itself can be modified; the planner may need to decompose an entire stack to access the block at the bottom. No such scenario exists in 3D scene graphs, as the support structure defined by receptacle and room layers stay constant throughout a planning problem, and only the objects' states are altered. Therefore, we argue that Blocks is an appropriate test bed for learning relational projective abstractions for planning problems, and can provide insight on how we might expect such a system to perform in the context of 3D scene graphs.

### 4.2.1 Sensitivity Analysis

In this sensitivity analysis, our goal is to quantify the effects of supervision quality on the relational inductive biases learned by the GNN in PLOI. Binary importance labels for all objects in the scene are acquired by compiling sufficient object sets, which can be *any* subset of objects such that *a planner* can solve still the problem; trivially, the largest sufficient object set is the full object set. During the data generation process, the size of the sufficient object set is iteratively reduced by sampling out objects randomly and attempting to replan. While the planner is not optimal, it serves its purpose as a validation tool for the stochastically generated object set. Thus, the sufficient object sets in PLOI always represents a conservative estimate of object importance, and training the GNN model with these sets will lead to conservative importance predictions.

Indeed, we can expect high recall rates upon a fine-tuned importance threshold, but the precision of the model remains a concern. Thus, if we would like to impose an upper bound on the number of interactable objects for a given task, thereby transforming the object importance scoring problem into one of object ranking, we might not expect the GNN to perform exceptionally well.

The PLOI model is trained on 40 problem instances for several epochs – training takes approximately 30 minutes on an Nvidia RTX 3090. Both the training and test problems contain on the order of a hundred objects, with 20-25 of those objects as part of the goal specification. The states of the remaining objects are redundant under the closed world assumption.

Table 7: Runtime of PLOI and Random Guidance in the Blocks domain.

| Problem Instance (no. objects) | Random Guidance (seconds) | PLOI (seconds) |
|---|---|---|
| **112 objects** | 20.034 | 0.298 |
| **126 objects** | 15.736 | 0.411 |
| **131 objects** | 16.323 | 0.531 |
| **135 objects** | 28.717 | 0.469 |
| **136 objects** | 29.170 | 0.524 |
| **136 objects** | 35.765 | 0.343 |
| **138 objects** | 44.388 | 0.411 |
| **139 objects** | 22.727 | 0.717 |
| **140 objects** | 45.021 | 0.312 |
| **152 objects** | 47.144 | 0.660 |
| **Mean ($\mu$)** | 20.503 | 0.468 |
| **Standard Deviation ($\sigma$)** | 11.412 | 0.135 |

We first compare PLOI to a Random Guidance module which scores the importance of objects as samples from a uniform random distribution; the results are presented in Table 7. The GNN performs an impressive two orders of magnitude faster than the Random Guidance module on average, exemplifying the high recall of the GNNs' predictions after being trained conservatively. Moreover, the runtime variance of PLOI is also two orders of magnitude smaller than that of Random Guidance. This indicates that leveraging planners to validate sufficient object sets offline suppresses much of the noise from the object sampling process, despite the suboptimalities of the planner. Thus, the planner enforces contextual independence of objects that have been sampled out of the object set during data generation. Random Guidance employs no such notion of contextual independence, and hence, it is equivalent to applying the offline data generation routine used in PLOI as a planner.

The above result should not come as a surprise, as the GNNs' capacity to learn useful relational inductive biases despite imperfections in training supervision and the validating planner is PLOI's main contribution. However, in real-world settings, having a robot accomplish a task with a minimal set of objects is not an unreasonable requirement. As we have seen in SGPlan, classical planners often exhibit greedy behaviours which involve objects and receptacles that need not be interacted with to solve the problem. If we task a robot with *cook dinner*, we may not want the robot to use all available cooking supplies in the process, even if they all nearby and have a reasonable affordance to the task. Instead, if the GNN was capable of sufficiently ranking objects, we could simply supply an extra task parameter which dictates the maximum number of objects that the plan should involve. However, ranking requires precise estimates of object importance, which is difficult to induce when training on noisy data. As we are unable to quantitatively verify precision against minimal object sets, we qualitatively gauge the precision of the GNN model in another experiment that attempts to plan with the top-K scored important objects. We sample top-K objects ranging from 25-135, as illustrated in Figure 19.

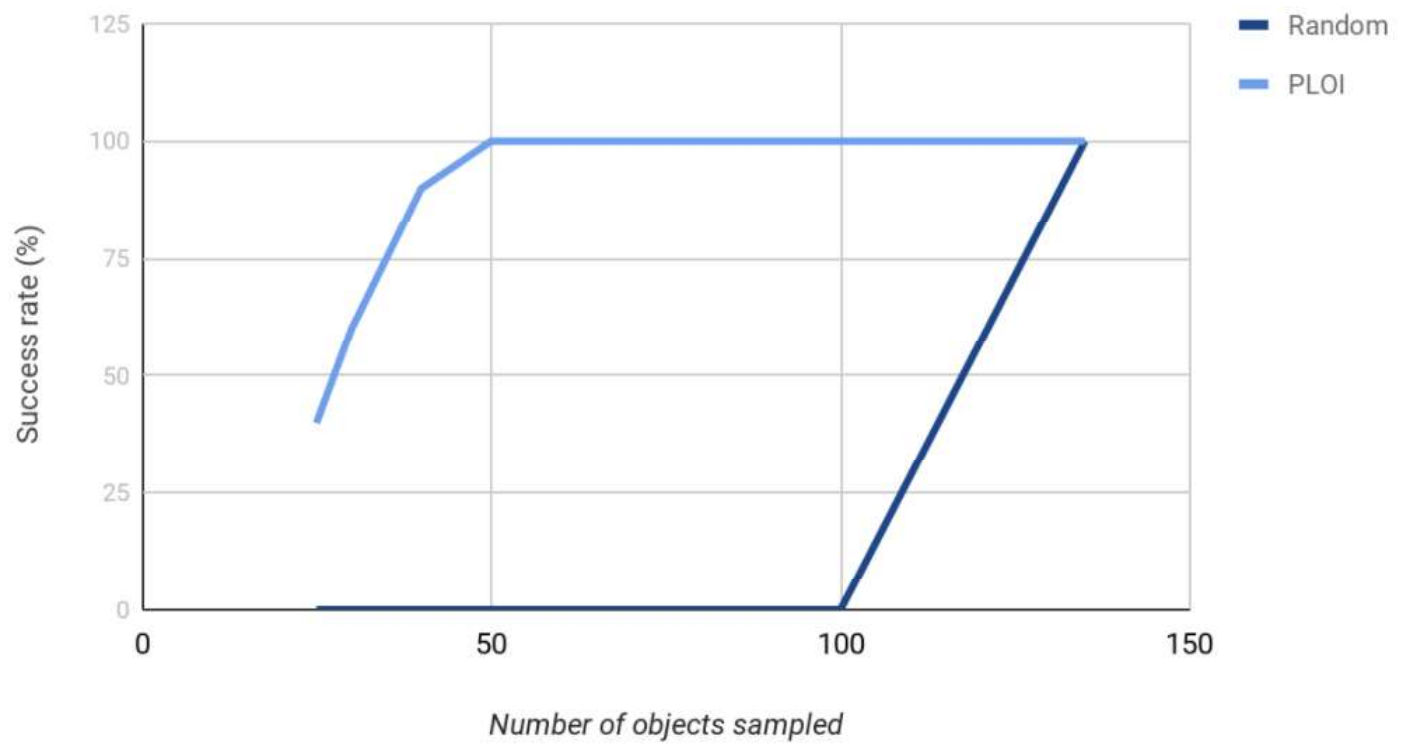

Figure 19: Success rate with respect to number of sampled objects – PLOI versus Random Guidance.

The analysis indicates that the GNN model is able to rank objects with reasonable precision. Task completion rates of 100% are achieved with the top-50 classified objects (135 total objects). The ability to solve problems with a 50% success rate while sampling only 30 objects is another interesting result, considering 20-25 of those objects are part of the goal condition, and interaction with other objects is likely required to achieve the goal. Random sampling proves unable to solve such problems with under 100 sampled objects.

It is possible that enriching the planning problem's relational representation with more informative node and edge attributes can lead to more precise object rankings. Hence, we append a distance metric to goal conditioned edges in Blocks. We hypothesize that the model will prioritize objects near large spatial edge attributes, as those regions of the domain are *further* from the goal state configuration, and hence, might require more interaction. The results that arise from the modification are discussed in Section 4.2.2.

Table 8: Effect of supervision quality on the GNN object importance scorer.

| **Problem Instance (no. objects)** | **Trained with Top 100 Objects (seconds)** | **Trained with Top 50 Objects (seconds)** |
|---|---|---|
| **126 objects** | 3.812 | 0.411 |
| **136 objects** | 2.536 | 0.343 |
| **138 objects** | 2.898 | 0.411 |
| **152 objects** | 5.336 | 0.660 |
| **Mean ($\mu$)** | 3.646 | 0.456 |
| **Standard Deviation ($\sigma$)** | 1.081 | 0.121 |

In a final sensitivity experiment, we uncover the dynamics between the size of the sufficient object set and the quality of predictions that can be expected from the GNN importance scorer. The results shown in Table 8 highlights a potential weak point of the PLOI system, where doubling the number of positive labels from 50 to 100 in the supervising sufficient object set produces runtimes that are approximately 8 times slower. Currently, the only strategy for obtaining smaller sufficient object sets is to re-run the stochastic sampling process in the hopes that the greedy search for important objects will terminate after more iterations. Given the clear limitations of this data generation procedure, and the strong correlations that exist between the size of sufficient objects sets and the inference quality of the GNN model, we believe that this component of PLOI is subject to improvement. As such, we demonstrate the results of our proposed regression planning based alternative to data generation in the following section.

### 4.2.2 Empirical Study of Proposed Models

In this section, we present the empirical results of our extensions to the PLOI planning system, as motivated in Section 3.2: (1) a graph attention network (GAT) variant for the prediction of object and relation importance; (2) leveraging supervision generated by regression planners (GAT-Regr); (3) incorporating spatial edge attributes in the relational representation of the Blocks domain (GAT-Spatial). The models are trained on a dataset comprised of 40 Blocks rearrangement problems, where each task contains between 10-15 objects. Evaluation is conducted on a test set of 10 complex problems with sizes that range between 15-55 objects.

Table 9: Model comparison on Blocks Test, including PLOI baselines.

| | FD [21] | RAND | K-NN | POLICY | ILP AG [10] | GNN AG [7] | PLOI [7] | PLOI 1-SHOT | GAT | GAT-REGR | GAT-SPATIAL |
|---|---|---|---|---|---|---|---|---|---|---|---|
| **TIME** | 7.47 | 49.99 | 0.52 | 7.25 | 2.33 | 52.95 | 0.34 | **0.25** | 0.34 | **0.21** | 0.32 |
| $\sigma_{time}$ | 0.07 | 15.80 | 0.06 | 0.77 | **0.04** | 27.68 | 0.11 | 0.05 | 0.11 | **0.01** | 0.12 |
| **FAIL** | 0.00 | 0.00 | 0.00 | 0.70 | 0.00 | 0.20 | 0.00 | 0.50 | 0.00 | 0.20 | 0.00 |

**Model Acronyms:** Object scoring based on k-nearest-neighbors to goal condition blocks (K-NN), action grounding via inductive logic programming (ILP AG [10]), custom graph neural network for action grounding (GNN AG [7]), our graph attention network (GAT), our model supervised by regression planners (GAT-Regr), our model with spatial edge attributes (GAT-Spatial).

Table 9 provides a summative comparison of all planning baselines evaluated in [7] and our proposed methods. In terms of planning efficiency, GAT-Regr outperforms all competitors. Because this model only attempts to plan once with the initial set of objects retained after thresholding, we also include a one-shot variant of PLOI trained with sufficient object sets generated by the Fast Downward progression planner. We observe that training with regression planner-based demonstrations yields a significant boost in both average plan time and inference robustness, as the one-shot object predictions of GAT-Regr contribute to a task completion rate of 80%, while the PLOI variant only solves half of the test problems. Furthermore, this model achieves impressive plan time consistency. As shown in Table 10, a majority of the object sets predicted by GAT-Regr facilitate planning within the 200-300 millisecond runtime bracket. In particular, for the ninth test problem, the resulting plan is an order of magnitude faster than all other models, while the fourth and fifth test problems approximately halve the competing runtimes. It is evident that the space of solutions covered by regression planners affords more effective scoring of objects. Intuitively, starting with the subset of objects in the goal specification and gradually increasing the object set throughout interactions in the backwards search should produce more concise sufficient object sets compared to the inverse process.

Table 10: Plan time comparison of proposed models on Blocks Test.

| MODEL | T1 | T2 | T3 | T4 | T5 | T6 | T7 | T8 | T9 | T10 | *μ* | *σ* |
|---|---|---|---|---|---|---|---|---|---|---|---|---|
| **PLOI [7]** | 0.20 | 0.26 | 0.34 | 0.50 | 0.41 | 0.37 | 0.55 | 0.22 | 0.22 | 0.35 | 0.34 | 0.11 |
| **PLOI 1-SHOT** | 0.20 | 0.26 | - | - | - | - | - | 0.22 | 0.22 | 0.35 | 0.25 | 0.05 |
| **GAT** | 0.20 | 0.26 | 0.34 | 0.50 | 0.41 | 0.37 | 0.55 | 0.22 | 0.22 | 0.35 | 0.34 | 0.11 |
| **GAT-REGR** | **0.17** | **0.22** | **0.24** | **0.24** | **0.27** | - | - | **0.22** | **0.01** | **0.31** | **0.21** | **0.01** |
| **GAT-SPATIAL** | 0.18 | 0.26 | 0.30 | 0.50 | 0.37 | **0.37** | **0.51** | 0.22 | 0.11 | 0.35 | 0.32 | 0.12 |

Table 11: Plan length comparison of proposed models on Blocks Test.

| MODEL | T1 | T2 | T3 | T4 | T5 | T6 | T7 | T8 | T9 | T10 | *μ* | *σ* |
|---|---|---|---|---|---|---|---|---|---|---|---|---|
| **PLOI [7]** | 34 | 28 | 52 | 64 | 56 | 52 | 64 | 26 | 28 | 52 | 45.6 | 14.31 |
| **PLOI 1-SHOT** | 34 | 28 | - | - | - | - | - | 26 | 28 | 52 | 33.6 | **9.58** |
| **GAT** | 34 | 28 | 52 | 64 | 56 | 52 | 64 | 26 | 28 | 52 | 45.6 | 14.31 |
| **GAT-REGR** | **26** | **24** | **30** | **32** | **36** | - | - | **26** | **5** | **48** | **28.38** | 11.38 |
| **GAT-SPATIAL** | 30 | 26 | 48 | 60 | 50 | **50** | **60** | 26 | 16 | 52 | 41.8 | 14.28 |

Further analysis on Tables 10-11 show that modifying the GNN model with edge attention and relation importance predictions (i.e., GAT [75]) performs equivalently to PLOI. Recall that the relationships between Blocks define the support structure of the domain. Since the entire support structure is described by edges of a single type, *on(?b1, ?b2)*, the expressiveness that GAT networks offers by attending to certain edge embeddings is almost negligible. The impact of relation importance predictions is also limited, since all of the action operators that govern state transitions in Blocks require the base support relations in order to pick and place objects. We anticipate that GNN architecture design will play a larger factor when applying this method to 3D scene graphs, which contain far more relationships and action operators with mutually exclusive preconditions.

Like the GAT-Regr model, spatial edge attributes (GAT-Spatial) led to improvements in planning efficiency and solution length over the baselines. However, the gains were of an incremental amount, as shown in Tables 10-11. These real-valued edge attributes differentiate edge embeddings throughout the planning problem, which in turn engages the effectiveness of edge attention and allows the GAT to learn more insightful importance representations. We expect this effect to be magnified in the feature richness of 3D scene graphs, perhaps requiring more expressive relational graph neural networks to differentiate interconnections (support structure, logical predicates) from intra-layer connections (entity-to-entity features).

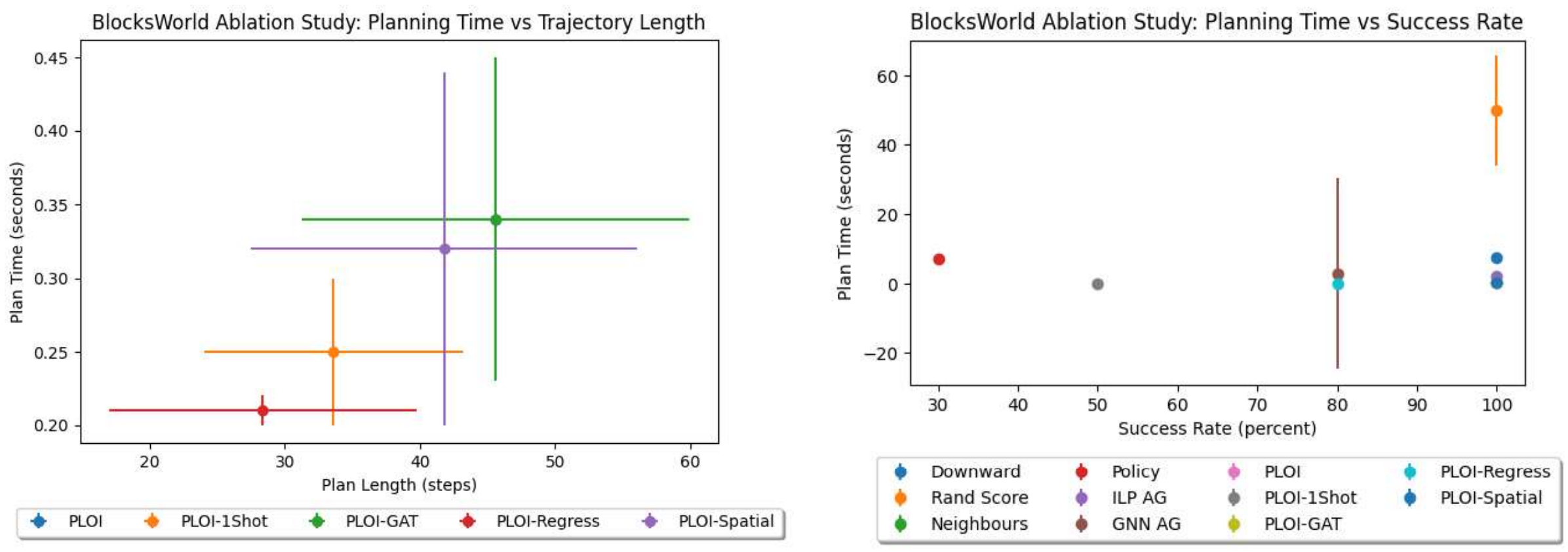

Figure 20: Ablation result – planning with projective abstractions. Plan time versus plan length (left) and plan time versus task completion (right). Error bars correspond to one standard deviation.

In two ablation studies shown in Figure 20, we compare the planners along multiple dimensions. In terms of plan time and plan length (left), we can clearly see the large performance margin achieved by GAT-Regr, instantiating regression planners as the most appropriate source of supervision for learning projective abstractions in future PLOI-type frameworks. GAT-Spatial proves a useful heuristic in the Blocks domain and has direct applications to 3D scene graphs. However, it yields the largest runtime variance of the ablated models. While GAT-Regr and GAT-Spatial display their respective benefits independently, we can expect to see further improvements by combining their complementing strengths. When comparing the task completion rate to plan times (right), GAT-Spatial stands out amongst the baselines. Unlike GAT-Regr, which we formulate as a one-shot model, GAT-Spatial is a complete planner which exponentially reduces the object threshold until a viable plan is found. As a result, we observe a 100% task completion rate with plan time progress made over PLOI due to the spatial edge attributes. Notice that the horizon of the test tasks proves too long for reinforcement learning policies to solve reliably. Moreover, while state-of-the-art lifted planners such as ILP AG [10] are comparable to PLOI models in the Blocks domain, they have been largely unsuccessful in a number of other classical planning domains [7].

Our findings on the topic of learning relational representations for efficient planning can be summarized as follows. First, regression planners show an outstanding capacity to generate concise and informative sufficient object sets to train the GNN classifier. The model trained with regression planner-based labels not only benefits in terms of precision, but also achieves higher recall rates, as inferred from 30% one-shot task completion margin achieved by GAT-Regr over PLOI one-shot (shown in Table 9). The benefits of edge attention become apparent when spatial edge attributes are incorporated into the relational input graph, as the GAT-Spatial model is able to distinguish long-range dependencies from short-range ones. This point is particularly relevant to 3D scene graphs, which contains spatial relations along with many other informative node and edge attributes. Altogether, we believe that projective abstractions for planning is a promising direction with many natural extensions to 3D scene graphs. We have already begun exploring these avenues and are excited of the results to come.

# Section 5: Conclusion

In this thesis, we explored the intricacies of current research at the intersection of three major topics: robot task planning, 3D scene graphs [4, 5], and Rearrangement [18]. We believe that the ability to create abstract plans is a fundamental characteristic of an intelligent agent. It allows for perplexing task specifications to be sequenced into subgoals that a robot is capable of enacting with simple motion primitives. However, planning requires an actionable state representation, which can used to simulate the effects of high-level actions into the future during the search for a satisficing solution. 3D Scene Graphs (SGs) provide such a representation in an object-centric relational graph structure, which describes the factored state of all interactable objects in scene. Their design supports a wide-range of realistic tasks in household and industrial applications, many of which can be regarded as instances of the Rearrangement challenge. However, this flexibility comes at the cost of planning efficiency. In particular, many of the objects and relationships defined in SGs are contextually irrelevant when conditioned on a task. Thus, while the design of systems that reliably construct SGs solely from visual-inertial data is an astounding accomplishment, it is important to test the hypothesis that these are the right actionable abstractions for long-horizon robot task planning.

We conducted our investigation along two objectives. Our first objective was to test the suitability of newly released embodied AI environments for scene graph task planning. This would provide valuable insight on the planning community's state of preparation for research at this niche. The results of our benchmarking efforts dictate the need for more a challenging task planning environment, as the combination of over-engineered tasks [106] in small-scale rooms [31] fail to present a significant challenge to the Fast Downward Planner [21]. In parallel, we aimed to develop a set of strategies to address the concern of planning efficiency in densely populated 3D scene graphs. In search for such solutions, we provided a holistic literature review of classical planning, modern planning, learning to search, and graph representation learning. We identified CAMPs [12] and PLOI [7] as an intriguing midpoint between pure planning and pure policy learning approaches. PLOI leveraged the inherent relational structure of planning problems with Graph Neural Networks to discern which objects are most relevant for a task. Several improvements were made over this framework with the incorporation of regression planners [13] and Graph Attention Networks [75].

To conclude, we believe that the release of a large-scale scene graph planning benchmark can contribute significantly to the vision and planning communities, and will foster research at the integration points between structured perceptual representations and robot task planning. As such, upcoming work holds the development of a benchmark built on Gibson [101], which offers the scale and diversity to sufficiently challenge both classical and learning-based planners. In addition, 3D Scene Graph [4] representations are openly accessible for most scenes in Gibson. We are concurrently working towards extensions on the concepts proposed by PLOI [7] and HMS [16] for application to 3D scene graphs. Our vision for the future of this field is one of excitement, particularly when we it grows to encompass motion planning and control, and reflects the complete formalism of Rearrangement. We hope to maintain our progress in the coming months, and contribute to this initiative.

# Appendix A. SGPlan - Additional Results

Table 12: Task completion rate ($TC_{\pi}$) of classical planners across SGPlan splits.

| | *Train* | *Validation* | *Test Seen* | *Test Unseen* |
|---|---|---|---|---|
| *FF-X* | 0.07 | 0.10 | 0.08 | 0.00 |
| *FD-Classic* | 1.00 | 1.00 | 1.00 | 1.00 |
| *FD-Dijkstra's* | 0.84 | 0.98 | 0.88 | 0.86 |
| *FD-CEA* | 1.00 | 1.00 | 1.00 | 1.00 |

Table 13: Average plan time ($T_{\pi}$) and plan length ($L_{\pi}$) of classical planners across SGPlan tasks.

| | *LoiL* | | *PaP* | | *SaP* | | *P2P* | | *PCoP* | | *PHeP* | | *PClP* | |
|---|---|---|---|---|---|---|---|---|---|---|---|---|---|---|
| | $T_{LoiL}$ | $L_{LoiL}$ | $T_{PaP}$ | $L_{PaP}$ | $T_{SaP}$ | $L_{SaP}$ | $T_{P2P}$ | $L_{P2P}$ | $T_{PCoP}$ | $L_{PCoP}$ | $T_{PHeP}$ | $L_{PHeP}$ | $T_{PClP}$ | $L_{PClP}$ |
| *FF-X* | 9.18 | 4.21 | 3.14 | 4.66 | - | - | 3.62 | 9.62 | - | - | - | - | 2.78 | 6.66 |
| *FD-Classic* | 0.39 | 4.67 | 0.56 | 5.18 | 0.74 | 5.0 | 0.51 | 9.37 | 1.13 | 8.09 | 1.27 | 8.53 | 0.98 | 7.59 |
| *FD-Dijkstra's* | 0.30 | 4.23 | 0.40 | 5.17 | 0.65 | 5.0 | 0.93 | 9.19 | 2.05 | 7.28 | 2.16 | 7.42 | 1.33 | 7.25 |
| *FD-CEA* | 0.31 | 5.25 | 0.39 | 5.24 | 0.66 | 5.0 | 0.41 | 9.62 | 0.87 | 8.36 | 0.81 | 8.86 | 0.71 | 7.82 |

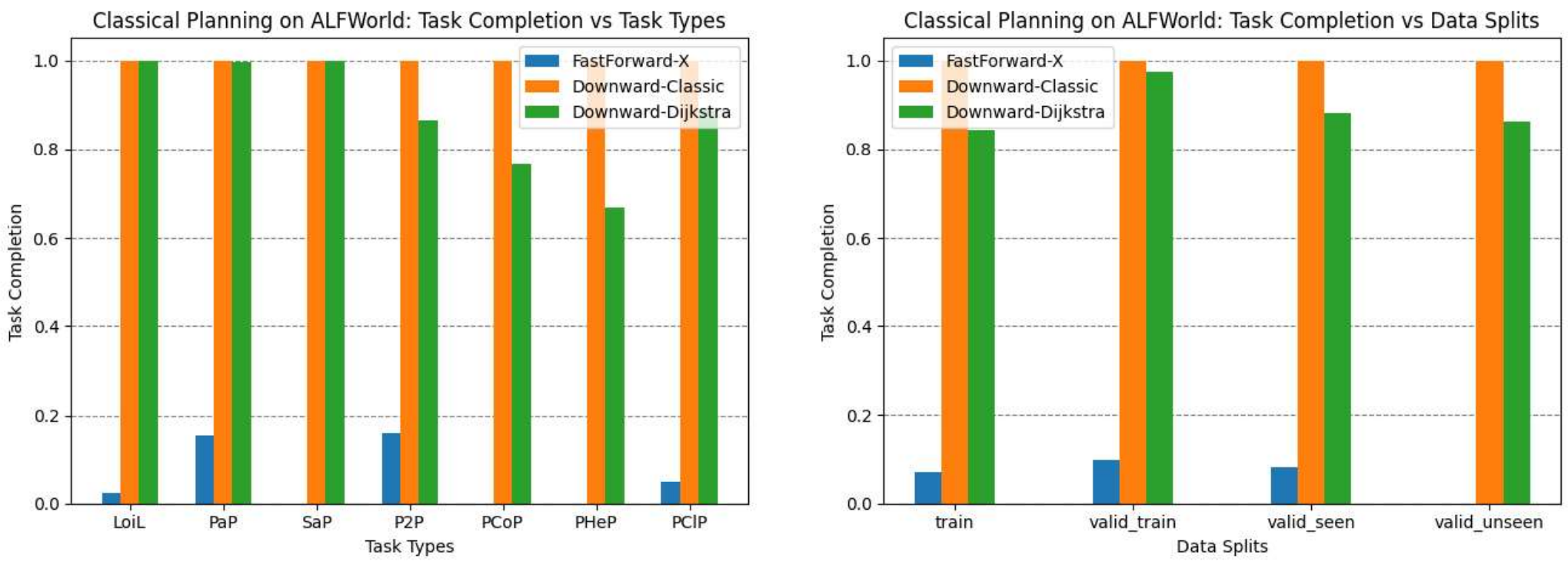

Figure 21: Task completion rate of classical planners across task types (left) and SGPlan splits (right).

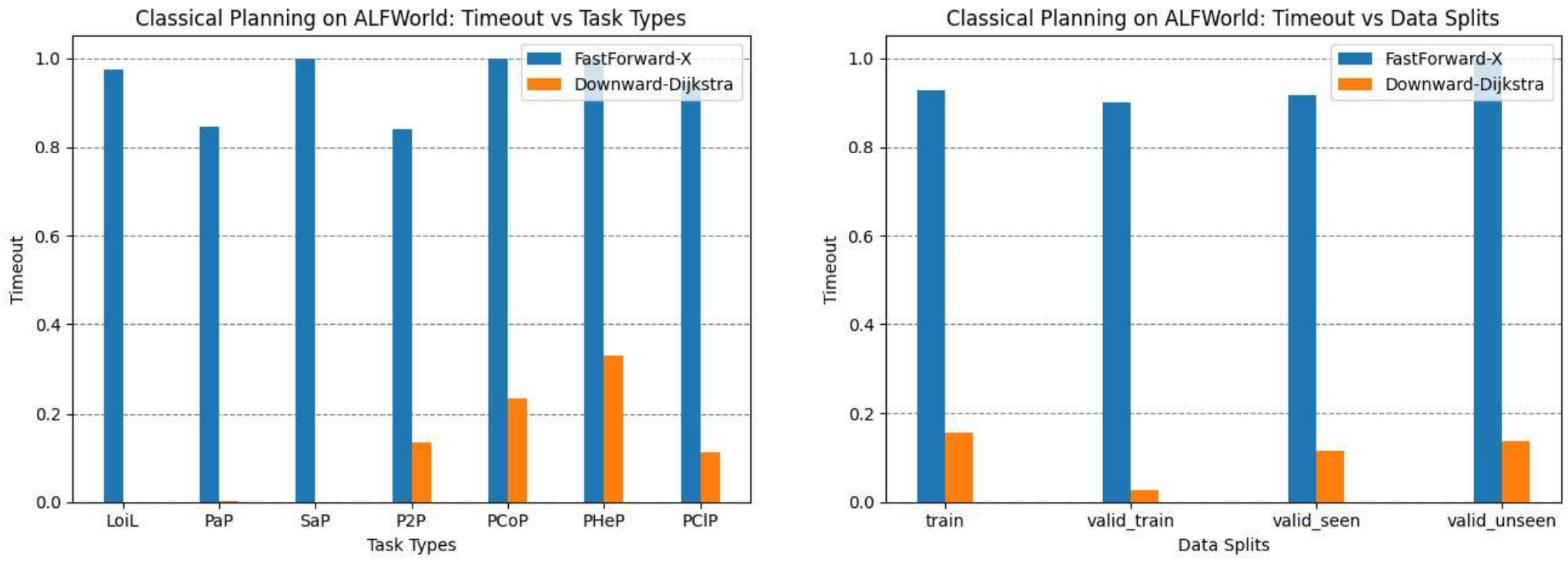

Figure 22: Ten second timeout rates of classical planners across task types (left) and SGPlan splits (right).

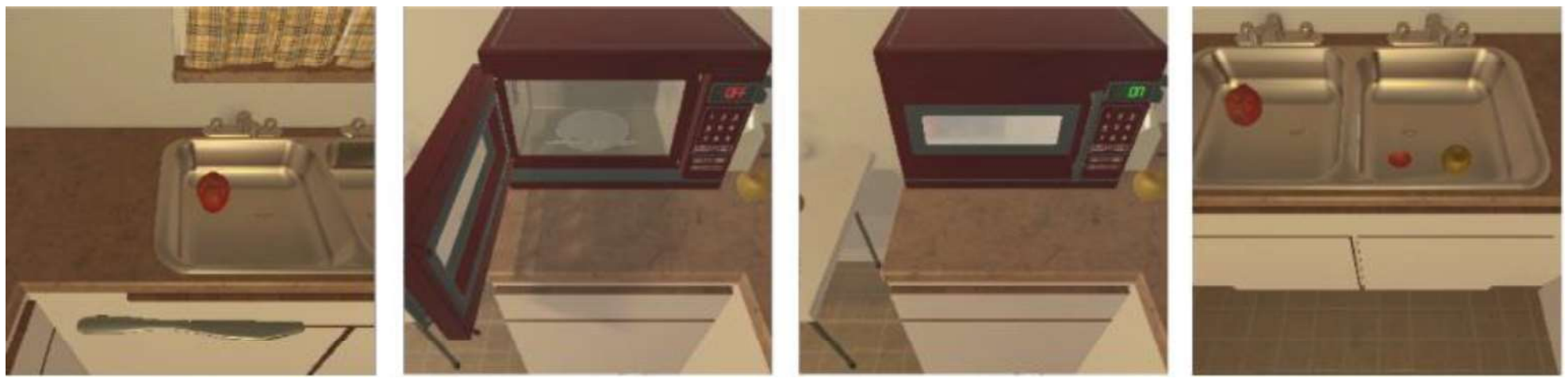

Figure 23: Unsafe trajectory generated by DiNo [9] on a pick-heat-and-place task.

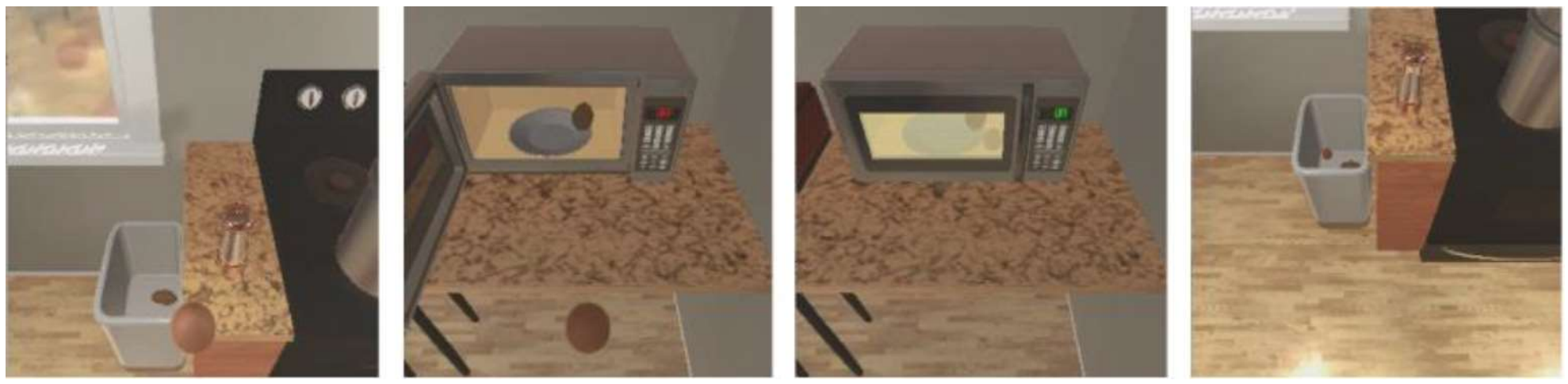

Figure 24: Unsafe trajectory generated by Fast Forward [20] on a pick-heat-and-place task.